\documentclass[9pt]{scrartcl}
\setkomafont{title}{\normalfont\bfseries}

\usepackage[bindingoffset=0.2in,%
left=0.5in,right=0.5in,top=1in,bottom=1in,%
footskip=.25in]{geometry}
\usepackage{tikz}
\usepackage{authblk}
\usepackage[ruled,vlined,linesnumbered,algo2e]{algorithm2e}
\usepackage{booktabs}
\usepackage{multirow}
\usepackage{array}
\usepackage{pdflscape}
\usepackage{amsmath}
\usepackage{floatrow}
\usepackage{rotating}
\usepackage{footnote}
\usepackage{amssymb,amsmath,amsthm}
\usepackage{mathtools}
\usepackage{breqn}
\usepackage{graphicx}
\usepackage[export]{adjustbox}
\usepackage[caption=false]{subfig}
\usepackage{tabularx}
\usepackage{amsfonts}
\usepackage{calc}
\usepackage{etoolbox}
\usepackage{calc}
\usepackage[bottom,hang]{footmisc}
\usepackage[hyphens]{url}
\KOMAoption{fontsize}{8.5pt}

\newcommand{\citet}[1]{\citeauthor{#1}~\shortcite{#1}}
\newcommand{\citep}{\cite}

\title{\Large Parallel Clustering of Single Cell Transcriptomic Data with Split-Merge Sampling on Dirichlet Process Mixtures}
\author[1]{\normalsize Tiehang Duan}
\author[2]{\normalsize Jos\'e P. Pinto}
\author[1]{\normalsize Xiaohui Xie \thanks{To whom correspondence should be addressed.}}
\affil[1]{\normalsize Department of Computer Science, University of California, Irvine, CA 92617, United States}
\affil[2]{\normalsize Systems Biology and Bioinformatics Laboratory (SysBioLab),  University
	of Algarve, Algarve, 8005-139, Portugal}

\date{}

\begin{document}

	\maketitle

\begin{abstract}	
\noindent
{\textbf{Motivation:} With the development of droplet based systems, massive single cell transcriptome data has become available, which enables analysis of cellular and molecular processes at single cell resolution and is instrumental to understanding many biological processes. While state-of-the-art clustering methods have been applied to the data, they face challenges in the following aspects: (1) the clustering quality still needs to be improved; (2) most models need prior knowledge on number of clusters, which is not always available; (3) there is a demand for faster computational speed.\\
\textbf{Results:} We propose to tackle these challenges with \underline{\textbf{Para}}llelized Split Merge Sampling on \underline{\textbf{D}}irichlet \underline{\textbf{P}}rocess \underline{\textbf{M}}ixture \underline{\textbf{M}}odel (the Para-DPMM model). Unlike classic DPMM methods that perform sampling on each single data point, the split merge mechanism samples on the cluster level, which significantly improves convergence and optimality of the result. The model is highly parallelized and can utilize the computing power of high performance computing (HPC) clusters, enabling massive inference on huge datasets. Experiment results show the model outperforms current widely used models in both clustering quality and computational speed. \\
\textbf{Availability:} Source code is publicly available on \url{https://github.com/tiehangd/Para_DPMM/tree/master/Para_DPMM_package}\\}
\end{abstract}

\section{Introduction}

Parallelized droplet based single cell transcriptomic profiling has achieved significant progress in recent years \cite{Zheng2017}. Compared to traditional methods, parallelized droplet based systems utilize Gel bead in EMulsion(GEM) to capture single cells in parallel (the co-occurrence of multiple cells in one GEM is eliminated by controlling the dilution in the reagent oil). The 3' messenger RNA digital counting is performed through the reading of unique molecular identifiers (UMI) in each GEM. Massive parallelized droplet based systems have the following properties: (1) Samples are processed in parallel in microfluidic chip with multiple channels, allowing the analysis of a much larger number of cells. (2) The multiplet rate (rate of multiple cells in one GEM) is controlled to be less than 2\% by limiting dilution, and performs direct counting of molecule copies using UMI. (3) The detection result of UMI is minimally affected by the composition of nucleobases and gene length, resulting in low transcript bias. Because of these properties, parallelized droplet based single cell transcriptomic profiling has resulted in the creation of mass single cell genomic datasets and lead to a number of advancements such as better approaches for transplant monitoring \cite{Athanasiadis2017} and detection of rare cell populations \cite{Proserpio2015}.

Cell clustering based on transcriptomic profiles plays an important role in single cell analysis. It identifies and characterizes cell subtypes from heterogeneous tissues and enhances understanding of cell identity and functionality. Classic clustering methods such as K-means \citep{Kmeans}, hierarchical clustering \citep{manning2008}, spectral clustering \citep{Ngonspectral} can be applied directly to single cell clustering. Given the high dimensionality of single cell data, a widely adopted approach involves combining dimension reduction with classic clustering.
Common combinations of methods include t-SNE with K-means \citep{Grn2015}, PCA with hierarchical
clustering \citep{zurauskiene2016} and Rt-SNE with model based clustering
\citep{Fraley2002, zurauskiene2016}.
  Other clustering methods tackle the dimensionality problem by replacing Euclidean distances with similarity measures that are robust in sparse high dimension space such as ranking on shared nearest neighbors (SNN) \citep{Satija2015}, ward linkage \citep{SINCERA2015}, and graph based clustering methods which perform graph partition by finding maximal cliques on the similarity matrix \citep{SNN2015}. 
  Dirichlet Mixture Model (DMM) is well suited for single cell clustering as the discrete counting information in the UMI matrix can be directly modeled through Multinomial distribution and conjugate prior likelihood pairs result in efficient inference \citep{Blei03}. 
  Recent applications of DMM to single cell analysis have achieved good results \citep{duVerle2016, Sun2017DIMMSCAD}. However, there are still challenges to be addressed:
  (1) There is demand for faster computational speed for newly created mass single cell datasets, which can be realized through parallelization and utilization of HPC clusters. However, standard DPMM methods are difficult to parallelize. (2) For challenging tasks, as shown in the experiment section, clustering quality can be significantly improved. (3) Most methods are designed for continuous data, while the scRNA-Seq data is formed of discrete UMI counts. Conversion of the UMI counts to continuous measure would alter the straight-forward interpretation and it is more appealing to directly model discrete data. (4) Most methods need prior knowledge on the number of clusters \citep{duVerle2016,xf2015}, which is not always available for rawly processed single cell data and limits their ability to identify cellular heterogeneity within the same cluster. 

\begin{figure*}[ht]
	\centering
	\includegraphics[width=0.78\textwidth]{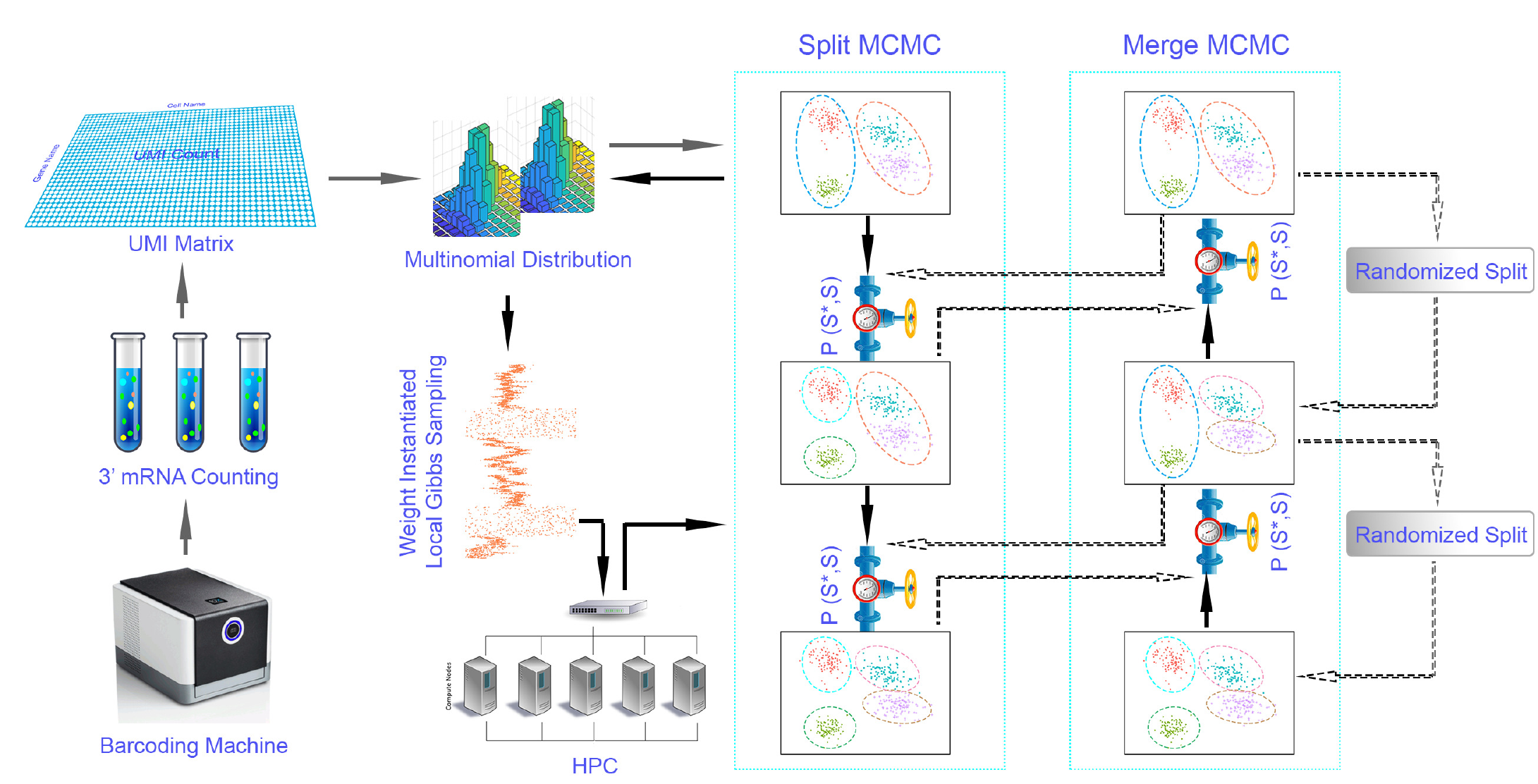}
	\caption{Workflow of Para-DPMM Model}
	\label{Figure 1}
\end{figure*}

The Para-DPMM model proposed in this paper addresses these limitations. Its inference is highly parallelized and can be readily implemented on large HPC clusters, which results in high computational speed. For large scaled datasets with tens of thousands of genes and cells, such as the fresh PBMC 68K dataset used in our second case study, the clustering is completed in a couple of minutes using 32 cores. The model is able to automatically determine the number of clusters with its nonparametric Bayesian setting. Its sampling is highly efficient. New clusters are created by splitting existing clusters instead of setting aside a single data point, which avoids going through the low probability density regions in the sampling space and achieves fast convergence and improved optimality. The model achieved more than 20\% improvement on ARI (adjusted rand index) for large challenging tasks over current widely used models in the experiment.

These improvements are due to a split-merge Markov Chain Monte Carlo (MCMC) inference algorithm that we developed for this problem. Unlike variational approximation \citep{blei2006,Kurihara2007} or collapsed Gibbs sampling \citep{Neal1992, Duan_2018_sequential}, the inference algorithm is a weight-instantiated sampling method, in which cluster parameters are explicitly instantiated as variables \citep{Ishwaran2001, Ishwaran2002}. Variational approximation algorithms lend themselves to parallelization, but are not guaranteed to converge to ground truth distribution. Collapsed Gibbs sampling enables intra-cluster parallelization \citep{Lovell2013arXiv,Williamson2013}, where the number of processes is parallelized to be of the same order as the number of clusters. Its parallelization level is relatively low. The split-merge sampling in Para-DPMM enables inter-cluster parallelization \citep{favaro2013, Papaspiliopoulos2008, ChangFisher2013}, in which threads running in parallel are of the same order as data points, resulting in a high level of parallelization. To improve sampling efficiency, new clusters are formed by either splitting an existing cluster or merging two clusters together. Local Gibbs sampling is performed inside each cluster to propose reasonable split proposals with high acceptance ratio.

\begin{table}[htbp]
	\centering
	\caption{Notations}
	\normalsize
	\begin{center}
		\scalebox{0.9}{
			\begin{tabular}{cl}
				\toprule
				Notation & Meaning \\
				\midrule
				$\vec{x}$ & collection of cells \\
				$\vec{c}$ & \multicolumn{1}{p{29em}}{cluster assignments of cells} \\
				$\vec{\theta}$ & cluster parameters \\
				$\pi$ & mixing proportions in the Dirichlet process \\
				$\lambda$ & Dirichlet hyper parameter for cluster parameters $\vec{\theta}$ \\
				$\alpha$ & parameter for Chinese restaurant process\\
				$c_{i}$ & cluster assignment for cell $i$ \\
				$\vec{\theta}_{k}$ & collection of parameters for the multinomial distribution in cluster $k$\\
				$\theta_{k}^{u}$ & parameter for multinomial distribution of gene $u$ in cluster $k$ \\
				$\vec{x}_{i}$ & the gene expression of $i$th cell \\
				$x_{i}^{u}$ & the UMI count of gene $u$ in cell $i$ \\
				$\vec{x}_{\{k\}}$ & the gene expression of cells assigned to cluster $k$ \\
				$\bar{c}$ & local split subcluster assignment \\
				$\bar{\theta}_{r}$ & parameters for local subclusters, $r \in \{0,1\}$ \\
				$n_{k}$ & number of cells in cluster $k$ \\
				$\bar{n}_{r}$ & number of cells in subcluster $r$, $r \in \{0,1\}$ \\
				$N$ & total number of cells \\
				$K$ & current number of clusters in the model \\
				$V$ & total number of genes \\
				\bottomrule
			\end{tabular}%
		}
	\end{center}
	\label{tab:addlabe6}%
\end{table}%

\section{Method}

\subsection{Data and Model Framework} The output of the droplet-based single cell profiling pipeline is a matrix storing UMI counts with rows indexing genes and columns indexing cells. Each entry in this UMI matrix  $x_{i}^{u}$ is the UMI count of gene $u$ barcoded in cell $i$. We use $\vec{x}_{i}$ to denote the expression of all genes in cell $i$ measured in terms of read counts. Single cell clustering is performed on the UMI matrix with size $V\times N$, where $V$ is the total number of genes and $N$ is the total number of cells. 

In the transcriptomic clustering model, the cluster assignment $c_{i}$ of cell $i$ is the discrete hidden variable to be inferred based on observed gene expression $\vec{x}_{i}$. The model is built on the Dirichlet process mixture model (DPMM), which is the infinite form of the Dirichlet mixture model (DMM). For detailed description of DPMM model please refer to \citep{dilan2010}. In the generative form of DPMM model, with parameters $\vec{\theta}_{k} \in \mathnormal{R}^{V}$, gene expression $\vec{x}_{i}$ is generated based on the Multinomial distribution 
\begin{equation} \label{eq:26}
p(\vec{x}_{i}|c_{i}=k,\vec{\theta}_{k})=\text{Multinomial}(\vec{x}_{i}|\vec{\theta}_{k})\sim\prod_{u=1}^{V}\theta_{k,u}^{x_{i}^{u}}
\end{equation}
where $\sum_{u=1}^{V}\theta_{k,u}=1$. Notation meaning is listed in Table \ref{tab:addlabe6}. Priors for $\vec{\theta}_{k}$ are accordingly set to be Dirichlet distribution with hyper parameter $\lambda$
\begin{equation} \label{eq:28}
\text{Dirichlet}(\vec{\theta}_{k}|\lambda)=\frac{\Gamma(\lambda V)}{\Gamma(\lambda)^{V}}\prod_{u=1}^{V}\theta_{k,u}^{\lambda-1}
\end{equation}
  For posterior inference of $c_{i}$ given gene expression $x_{i}$, the iterative inference process can be described as  
\begin{align} \label{eq:4}
(\pi_{1},...,\pi_{K},\pi_{K+1})&\sim p(\pi | \vec{c}, \alpha)\\
\vec{\theta}_{k} \propto p(\vec{x}_{\{k\}}|\vec{\theta}_{k})p(\vec{\theta}_{k}|\lambda)&\,\,\,\, \forall k\in \{1,...,K,K+1\}\\
c_{i} \propto p(c_{i}|\pi)p(\vec{x}_{i}|c_{i}&=k,\vec{\theta}_{k})
\end{align}
where $\{\pi_{1},...,\pi_{K}\}$ represents the mixing proportions of existing clusters and $\pi_{K+1}$ represents the proportion of next new cluster to be generated.

\subsection{Efficient Parallel Sampling for the DPMM Model}

 Implementing parallel inference for the DPMM model is not trivial. Careful examination of the dependence relationships among the variables is necessary. While collapse Gibbs sampling \citep{neal_2000_mc} simplifies the sampling process (when priors are conjugate to the likelihood), its parallelization is not straight forward \citep{ChangFisher2013} as data points become directly dependent on each other after the cluster parameters are integrated out. The cluster indicators $\vec{c}$ can be seen as a fully connected Markov Random Field (MRF) and can't be parallelized based on proofs in \citep{Gonzalez_parallelgibbs}.

For the split merge sampling adopted in this paper, the cluster parameters $\vec{\theta}$ are explicitly instantiated as variables.  The cluster assignments $\vec{c}$ and cluster parameters $\vec{\theta}$ can be mapped to a two coloring MRF with one color being $\vec{c}$ and the other being $\vec{\theta}$. Based on theorems in \citep{Gonzalez_parallelgibbs}, all cluster assignments $\vec{c}$ can then be sampled in parallel, as they are conditionally independent of each other given $\vec{\theta}$. Theoretically, the maximum number of computing cores that can be utilized in parallel equals the number of data points.

Sampling is inefficient in this naive parallel approach. It is difficult to open new clusters as parameters sampled directly from the prior are usually a poor fit of the data. Also, extremely large number of sampling steps are needed for common scenarios such as: (1) dividing the current cluster into more fine grained clusters; (2) transferring a significant portion of data points in the current cluster to another cluster; (3) merging two clusters. The naive approach has to go through a series of low probabilistic density intermediate steps in the sampling space to reach the more optimized setting. In real world applications where sampling time is limited, this approach leads to sub-optimality.

The split merge sampling mechanism was adopted to solve this problem. New clusters are created by splitting existing clusters, instead of setting aside a single data point. This endows newly created clusters with sensible parameters and data membership from the very beginning, and avoids going through low probability intermediate states, thus leading to faster convergence. To guarantee that the process converges to the desired stationary state, a MCMC is built to satisfy the detailed balance by either accepting or rejecting the splitting proposal. Merge moves are introduced to make the Markov chain ergodic, its proposal is accepted based on a separate acceptance ratio.

\subsection{Inference through Split/Merge MCMC Sampling}

The MCMC sampler is characterized by the states and acceptance ratio of state transitions. For the Para-DPMM model, each state is defined as $S=\{\vec{\pi}, \vec{\theta}, \vec{c}, \vec{x}\}$. For each update, the algorithm proposes a new state $S_{*}=\{\vec{\pi}_{*}, \vec{\theta}_{*}, \vec{c}_{*}, \vec{x}_{*}\}$ which is reachable from the old state by either a split or merge move. As the derivation for the two moves are similar, here we take split move as example. The proposed state is either accepted or rejected based on the acceptance ratio:

\begin{equation} \label{eq:7}
p(S_{*},S)=\text{min}\left[1, \frac{p(S_{*})}{p(S)}\frac{q(S|S_{*})}{q(S_{*}|S)}\right]
\end{equation}

where $p(S)$ is the likelihood of the old state, $p(S_{*})$ is the likelihood of the new state, $q(S_{*}|S)$ is the transition probability from old state to new state, and $q(S|S_{*})$ is the reversed transition probability. Updates with this acceptance ratio satisfy the detailed balance of Markov chain and are guaranteed to converge to the stationary state. 

 Derivation of the acceptance ratio is based on the specific split merge mechanism we choose. The random split with binomial distribution is straight forward, yet its performance is not satisfactory, as it doesn't utilize any information in the data points and the proposals are unlikely to be reasonable. The acceptance ratio is usually very low in this scenario.

An improved method is to run local Gibbs sampling in each cluster to learn cluster substructures before the split proposal. An additional indicator variable $\bar{c}=\{ 0, 1 \}$ is assigned to each data point in cluster $k$ to denote which data points will be in the sub-clusters after the possible split. Local Gibbs sampling computes the probability of assigning data points to either side of the split:
\begin{equation} \label{eq:8}
p(\bar{c}_{i}=r|\bar{c}_{\{r\},\neg i}, \vec{x}_{\{r\}}, \bar{\theta})=\frac{\bar{n}_{\{r\},\neg i} p(\vec{x}_{i}|\bar{\theta}_{r}, \bar{c}_{i}=r)}{\bar{n}_{\{0\},\neg i} p(x_{i}|\bar{\theta}_{0}, \bar{c}_{i}=0)+\bar{n}_{\{1\},\neg i}p(\vec{x}_{i}|\bar{\theta}_{1},\bar{c}_{i}=1)} \,\,\,\, \forall r\in \{0,1\}
\end{equation}
where $\bar{c}_{\{r\},\neg i}$ are the assignments to subcluster $r$ excluding cell $i$ and $\bar{n}_{\{r\},\neg i}$ is the number of cells in subcluster $r$ excluding cell $i$. Parameters for local subclusters are then updated based on
\begin{equation} \label{eq:19}
\bar{\theta}_{r} \propto p(\vec{x}_{\{r\}}|\bar{\theta}_{r})p(\bar{\theta}_{r}|\bar{\lambda})\,\,\,\, \forall r\in \{0,1\}
\end{equation}
where $\bar{\lambda}$ is the Dirichlet hyper parameter for subcluster parameters $\bar{\theta}$.

The number of iterations for local Gibbs sampling involves a trade off between accuracy and computational cost. In practice we found one iteration is already enough for the model to achieve decent performance. Transition probability $q(S^{*}|S)$ based on the local Gibbs sampling is a product of conditional probabilities of assigning each observation $i \in \{k\}$ to a split mixture component as given by Eq. (\ref{eq:8}). The transition probability from the new state back to old state $q(S|S^{*})$ is also needed. This reverse transition is the merge operation. In contrast to the split operation which has diversified splitting choices, the merge operation is deterministic as there is only one way to merge two components into one component, so $q(S|S^{*})=1$. 

 To calculate the acceptance ratio in Eq. (\ref{eq:7}), we also need to evaluate the ratio of likelihood between the new state and the old state $\frac{p(S^{*})}{p(S)}$. According to the generative procedure of DPMM, $\frac{p(S^{*})}{p(S)}$ can be decomposed as
\begin{equation} \label{eq:9}
\frac{p(S^{*})}{p(S)}=\frac{p(\vec{\pi}_{*}, \vec{\theta}_{*}, \vec{c}_{*}, \vec{x}_{*})}{p(\vec{\pi}, \vec{\theta}, \vec{c}, \vec{x})}=\frac{p(\vec{\pi}_{*})}{p(\vec{\pi})}\frac{p(\vec{c}_{*}|\vec{\pi}_{*})}{p(\vec{c}|\vec{\pi})}\frac{p(\vec{\theta}_{*}|\lambda)}{p(\vec{\theta}|\lambda)}\frac{p(\vec{x}|\vec{c}_{*},\vec{\theta}_{*})}{p(\vec{x}|\vec{c},\vec{\theta})}
\end{equation}
 $\frac{p(S^{*})}{p(S)}$ can be readily derived from Eq. (\ref{eq:9}) to be
 \begin{equation} \label{eq:18}
 \frac{p(S^{*})}{p(S)}=\alpha \frac{\pi_{k_{0}}^{\bar{n}_{k_{0}}-1}\pi_{k_{1}}^{\bar{n}_{k_{1}}-1}}{\pi_{k}^{n_{k}-1}}\frac{\Gamma(\lambda V)}{\Gamma(\lambda)^{V}}\frac{\prod_{u=1}^{V}\theta_{k_{0},u}^{\lambda-1}\prod_{u=1}^{V}\theta_{k_{1},u}^{\lambda-1}}{\prod_{u=1}^{V}\theta_{k,u}^{\lambda-1}}\times\frac{\Big(\prod_{i\in \{k_{0}\}}\prod_{u=1}^{V}\theta_{k_{0},u}^{x_{i}^{u}}\Big)\Big(\prod_{i\in \{k_{1}\}}\prod_{u=1}^{V}\theta_{k_{1},u}^{x_{i}^{u}}\Big)}{\Big(\prod_{i\in \{k\}}\prod_{u=1}^{V}\theta_{k,u}^{x_{i}^{u}}\Big)}
 \end{equation}
The detailed derivation is included in the supplementary material.

\subsection{Random Splits in Merge Moves}

A key consideration when constructing the MCMC sampler is to avoid the acceptance rate to be too small. For this reason, as mentioned in the previous section, we replaced random split with local Gibbs sampling when designing split moves. When the split is more reasonable, the likelihood of the new state $p(S^{*})$ significantly increases, thus increasing the acceptance rate. Merge moves can be seen as split moves going from the new state back to the old state. To increase the acceptance rate of merge moves, we should do exactly the opposite. And we included in the model a separate pair of merge/split moves which is randomized to propose good merges (as here the splitted cluster is the old state whose likelihood we are trying to decrease). For randomized merge moves, as $p(\bar{c}_{i}=r|\bar{c}_{\{r\},\neg i}, \vec{x}_{\{r\}})$ is simply $\frac{1}{2}$, the ratio of transition probability becomes
\begin{equation} \label{eq:23}
\frac{q(S|S_{*})}{q(S_{*}|S)}=\bigg(\frac{1}{2}\bigg)^{n_{k_{0}}+n_{k_{1}}-2}
\end{equation}
The derivation of $\frac{p(S_{*})}{p(S)}$ is similar to the split move.

Please note the split moves and merge moves that take place in the model belong to two independent MCMC chains. The integrated dynamic process thus formed is a rational MCMC with guaranteed convergence as long as the atomic moves are selected randomly from the two chains and each of the chains satisfies detailed balance \citep{tierney1994}.

\section{Performance in Cellular Heterogeneity Analysis} The Para-DPMM model was applied to the challenging task of distinguishing three T cell types (CD4+/CD25+ regulatory T cells, CD4+/CD45RA+/CD25- naive T cells and CD8+/CD45RA+ naive cytotoxic T cells) similar to \citep{Sun2017DIMMSCAD}. The data was provided by 10X Genomics and is publicly available (\citep{Zheng2017}). Three data sets of different scales were used: (1) a set of 1200 cells with the 1000 top variable genes (small scale, referred to as S-Set below), (2) a set of 3000 cells with the 3000 top variable genes (medium scale, referred to as M-Set) and (3) a set of 6000 cells with the 5000 top variable genes (large scale, referred to as L-Set). In these datasets, cells were randomly selected from the population, we ensured that each cell type was equally represented in the datasets. The top variable genes were selected based on their standard deviations across the cell transcriptome profiles in the UMI matrix.

We compared Para-DPMM's performance with other currently widely used models, including Seurat (\citep{Satija2015}), CellTree (\citep{duVerle2016}), PCA-Reduce (\citep{zurauskiene2016}), SC3 (\citep{Kiselev_2017}), SIMLR (\citep{Wang_2017}), CIDR (\citep{Lin_2017}) and DIMM-SC (\citep{Sun2017DIMMSCAD}). For models needing prior knowledge on the number of clusters, we set it to the ground truth value.
The results are shown in Table \ref{label1}. The model's performance was measured with three benchmarks: Adjusted Rand Index (ARI), Rand Index (RI) and Hubert's Index (HI). Rand Index (RI) measures the similarity between two clusterings, it ranges between 0 and 1 with a perfect match being scored 1. Adjusted Rand Index (ARI) is the corrected-for-chance version of Rand Index, it scores 0 for random matches. Hubert's Index (HI) (\citep{Hubert1985}) is another popular metric for comparing partitions. It has the advantage of probabilistic interpretation in addition to being corrected for chance. Its value ranges between -1 and 1. The analysis below mainly refers to ARI due to its wide adoption in the field.

As shown in Table \ref{label1}, Para-DPMM outperformed all comparison methods for a large margin on all experiment settings, and the trend is more significant in the large data setting (L-Set), where it achieved approximately 5\% improvement on ARI compared to SC3, and is more than 20\% better than the other comparison methods. We further applied Para-DPMM to the full dataset, which includes 32,695 cells and 32,738 genes, where the model achieved a 71.47\% score on ARI.

\begin{table*}[htbp]
	\centering
	\caption{Performance Comparison on Different Data Scales}
	\begin{center}
		\scalebox{0.73}{
			\begin{tabular}{l|rrr|rrr|rrr}
				\toprule
				& \multicolumn{3}{c|}{S-Set}    & \multicolumn{3}{c|}{M-Set}    & \multicolumn{3}{c}{L-Set} \\
				\midrule
				& \multicolumn{1}{c}{ARI} & \multicolumn{1}{c}{RI} & \multicolumn{1}{c}{HI} & \multicolumn{1}{c}{ARI} & \multicolumn{1}{c}{RI} & \multicolumn{1}{c}{HI} & \multicolumn{1}{c}{ARI} & \multicolumn{1}{c}{RI} & \multicolumn{1}{c}{HI} \\
				\midrule
				Para-DPMM & \textbf{0.654 $\pm$ 0.021} & \textbf{0.849 $\pm$ 0.011} & \textbf{0.699 $\pm$ 0.023} & \textbf{0.670 $\pm$ 0.012}  & \textbf{0.855 $\pm$ 0.004} & \textbf{0.711 $\pm$ 0.008} & \textbf{0.688 $\pm$ 0.016} & \textbf{0.863 $\pm$ 0.008} & \textbf{0.726 $\pm$ 0.016} \\
				DIMM-SC & 0.578 $\pm$ 0.029 & 0.803 $\pm$ 0.006 & 0.606 $\pm$ 0.012 & 0.352 $\pm$ 0.009 & 0.662 $\pm$ 0.018 & 0.324 $\pm$ 0.036 & 0.331 $\pm$ 0.013 & 0.650 $\pm$ 0.023 & 0.301 $\pm$ 0.047 \\
				CellTree & 0.270 $\pm$ 0.006 & 0.637 $\pm$ 0.015 & 0.274 $\pm$ 0.031 & 0.289 $\pm$ 0.009 & 0.643 $\pm$ 0.016 & 0.285 $\pm$ 0.032 & 0.273 $\pm$ 0.008 & 0.634 $\pm$ 0.024 & 0.268 $\pm$ 0.048 \\
				Seurat & 0.503 $\pm$ 0.017 & 0.776 $\pm$ 0.010 & 0.553 $\pm$ 0.019 & 0.576 $\pm$ 0.032 & 0.815 $\pm$ 0.008 & 0.630 $\pm$ 0.015 & 0.463 $\pm$ 0.028 & 0.785 $\pm$ 0.018 & 0.569 $\pm$ 0.036 \\
				PCA-Reduce & 0.294 $\pm$ 0.015 & 0.684 $\pm$ 0.018 & 0.368 $\pm$ 0.036 & 0.284 $\pm$ 0.016 & 0.681 $\pm$ 0.021 & 0.363 $\pm$ 0.041 & 0.302 $\pm$ 0.014 & 0.688 $\pm$ 0.016 & 0.376 $\pm$ 0.032 \\
				K-means & 0.312 $\pm$ 0.014 & 0.680 $\pm$ 0.004 & 0.360 $\pm$ 0.008 & 0.302 $\pm$ 0.007 & 0.678 $\pm$ 0.012 & 0.355 $\pm$ 0.023 & 0.312 $\pm$ 0.019 & 0.683 $\pm$ 0.005 & 0.367 $\pm$ 0.010 \\
				SC3 & 0.602 $\pm$ 0.018 & 0.823 $\pm$ 0.006 & 0.646 $\pm$ 0.012 & 0.614 $\pm$ 0.026 & 0.828 $\pm$ 0.018 & 0.657 $\pm$ 0.036 & 0.640 $\pm$ 0.017 & 0.840 $\pm$ 0.010 & 0.680 $\pm$ 0.020 \\
				SIMLR & 0.203 $\pm$ 0.014 & 0.606 $\pm$ 0.006 & 0.212 $\pm$ 0.012 & 0.334 $\pm$ 0.011 & 0.699 $\pm$ 0.013 & 0.398 $\pm$ 0.026 & 0.381 $\pm$ 0.008 & 0.724 $\pm$ 0.012 & 0.449 $\pm$ 0.024 \\
				CIDR & 0.222 $\pm$ 0.011 & 0.605 $\pm$ 0.014 & 0.209 $\pm$ 0.028 & 0.196 $\pm$ 0.009 & 0.617 $\pm$ 0.015 & 0.235 $\pm$ 0.030 & 0.205 $\pm$ 0.016 & 0.628 $\pm$ 0.009 & 0.255 $\pm$ 0.018 \\
				\bottomrule
			\end{tabular}%
		}
	\end{center}
	\label{label1}%
\end{table*}%

As mentioned in the previous section, the performance improvement is due to the split merge mechanism which enables the model to make efficient moves in the sampling space and avoid being trapped in sub-optimal situations. The underlying Dirichlet Process allows the model to automatically decide the most appropriate number of clusters for the data, and the parallelized sampling enhances the convergence speed.

\begin{figure*}[ht]
	\centering
	\subfloat[]
	{
		\includegraphics[scale=.25,valign=t]{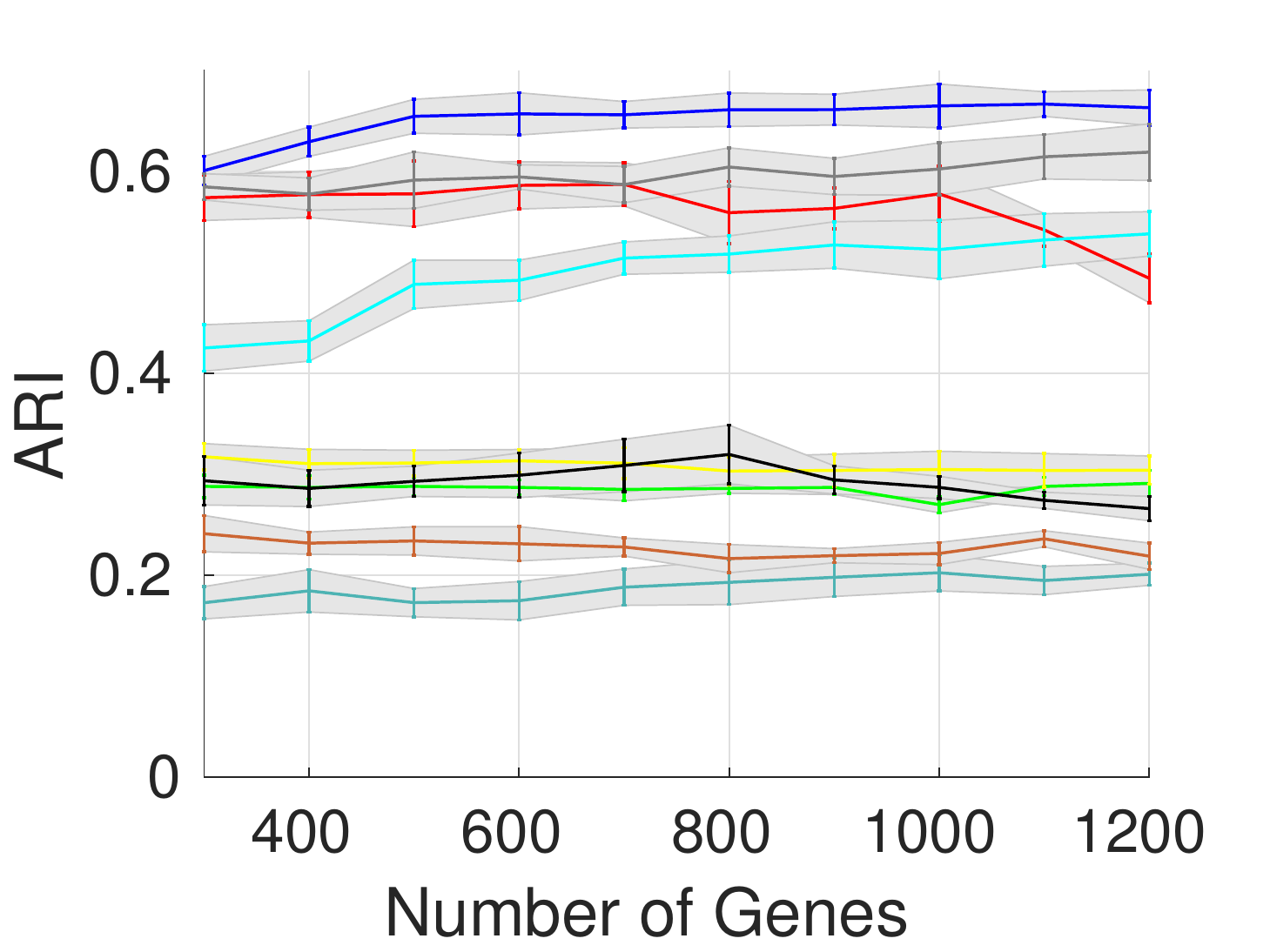}
		\label{Fig9_1}
	}
	\subfloat[]
	{
		\includegraphics[scale=.25,valign=t]{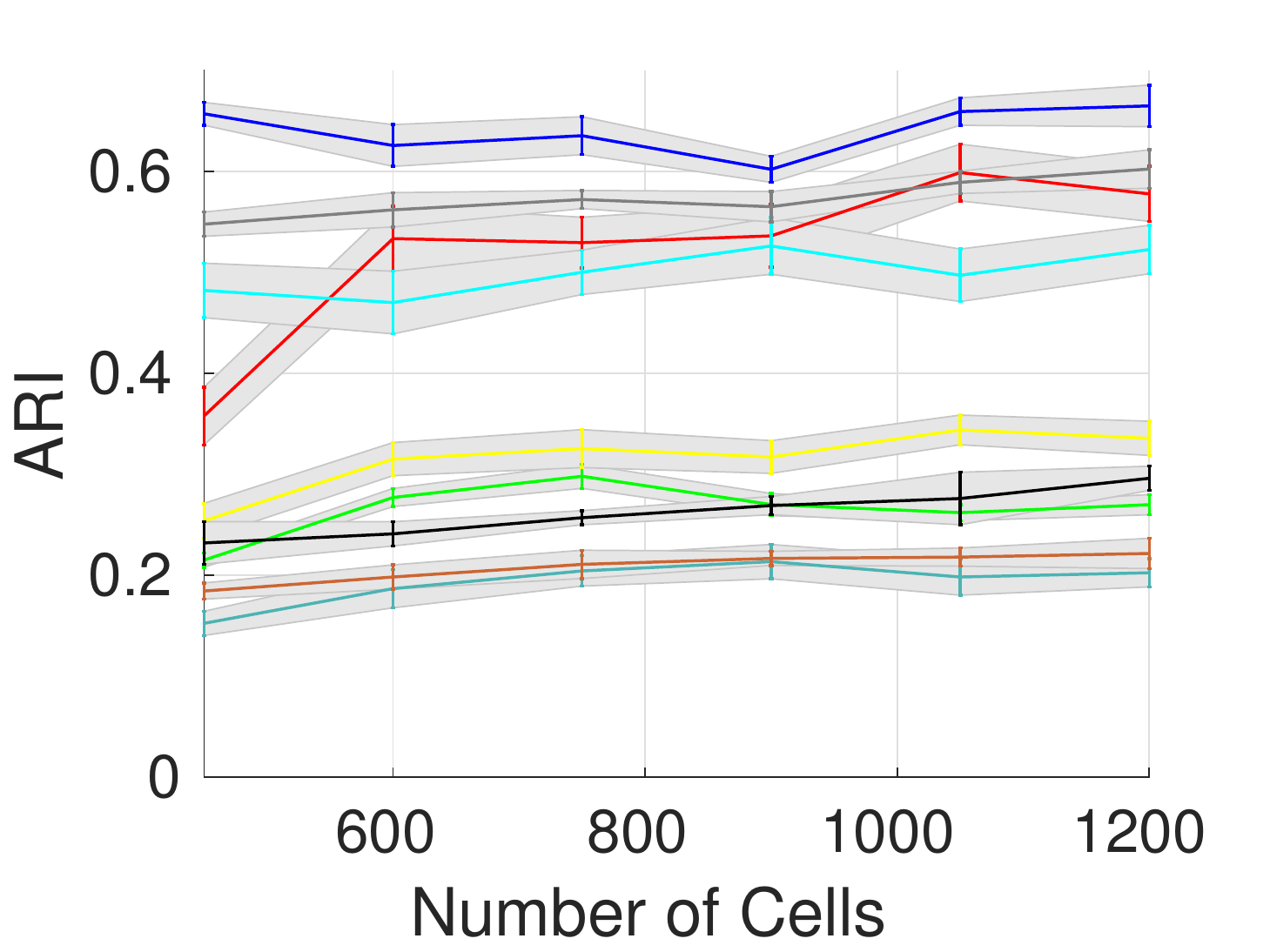}
		\label{Fig9_2}
	}
	\subfloat[]
	{
		\includegraphics[scale=.25,valign=t]{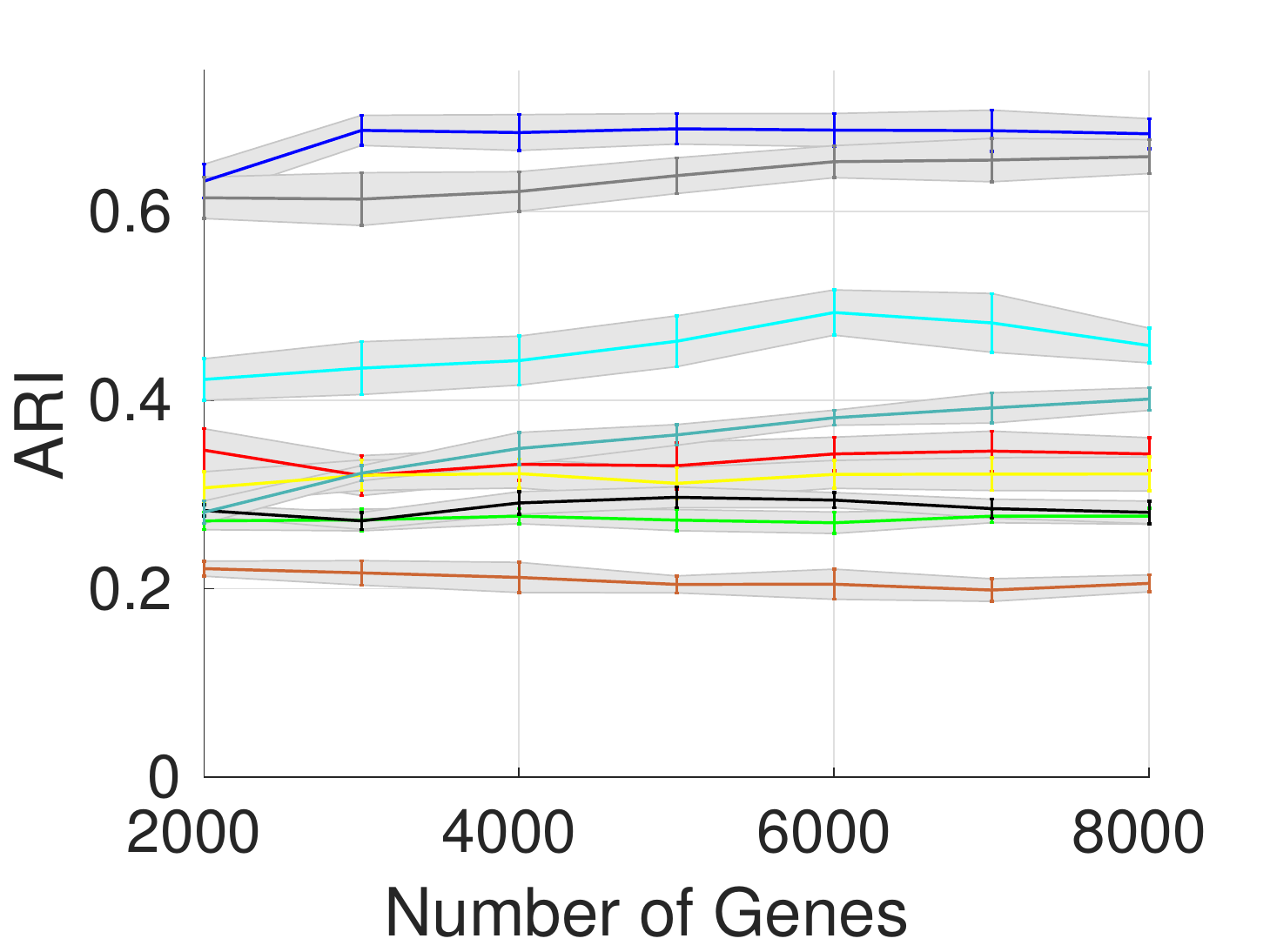}
		\label{Fig9_3}
	}
	\subfloat[]
	{
		
		\includegraphics[scale=.265,valign=t]{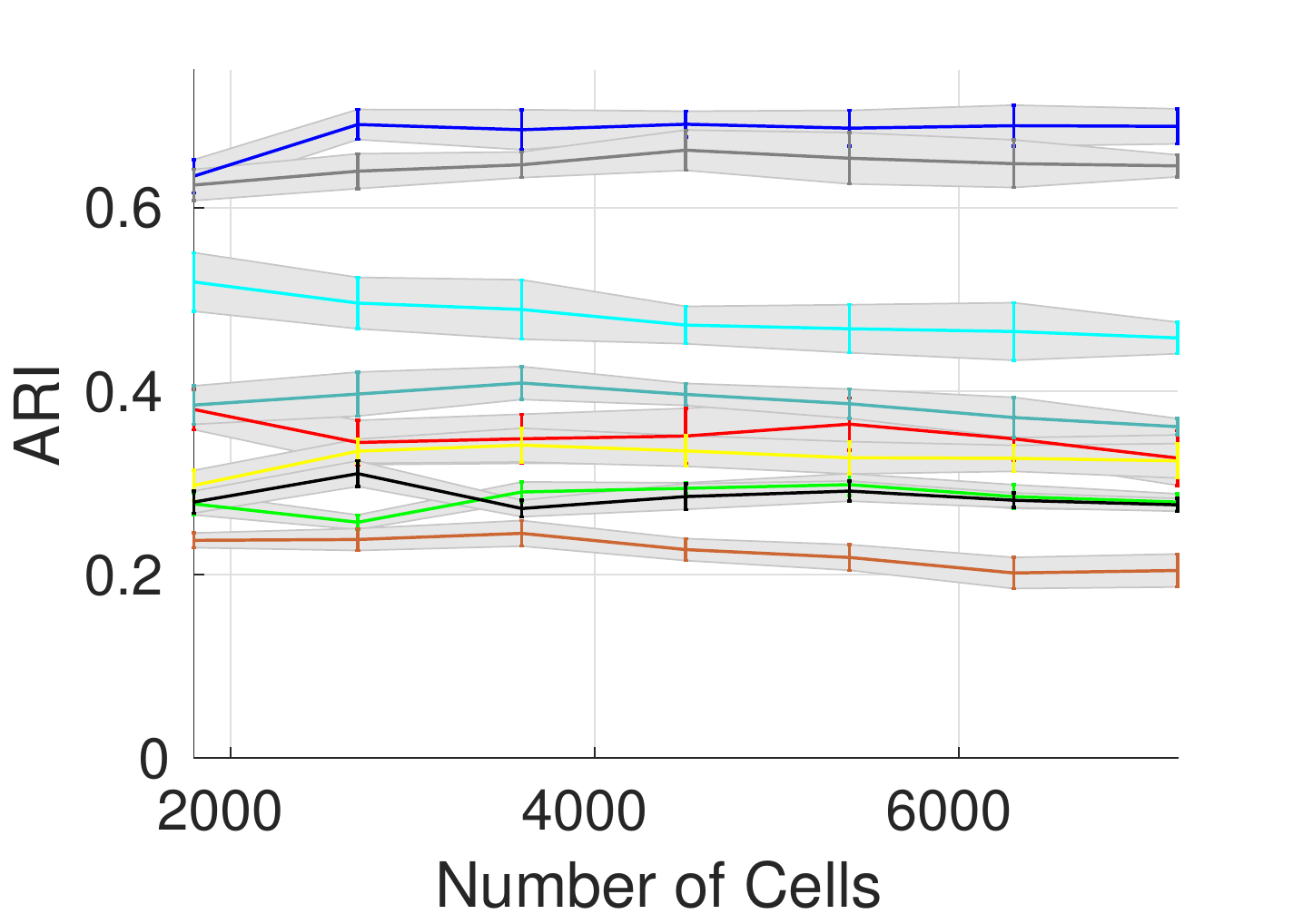}
		\includegraphics[scale=.15,valign=t]{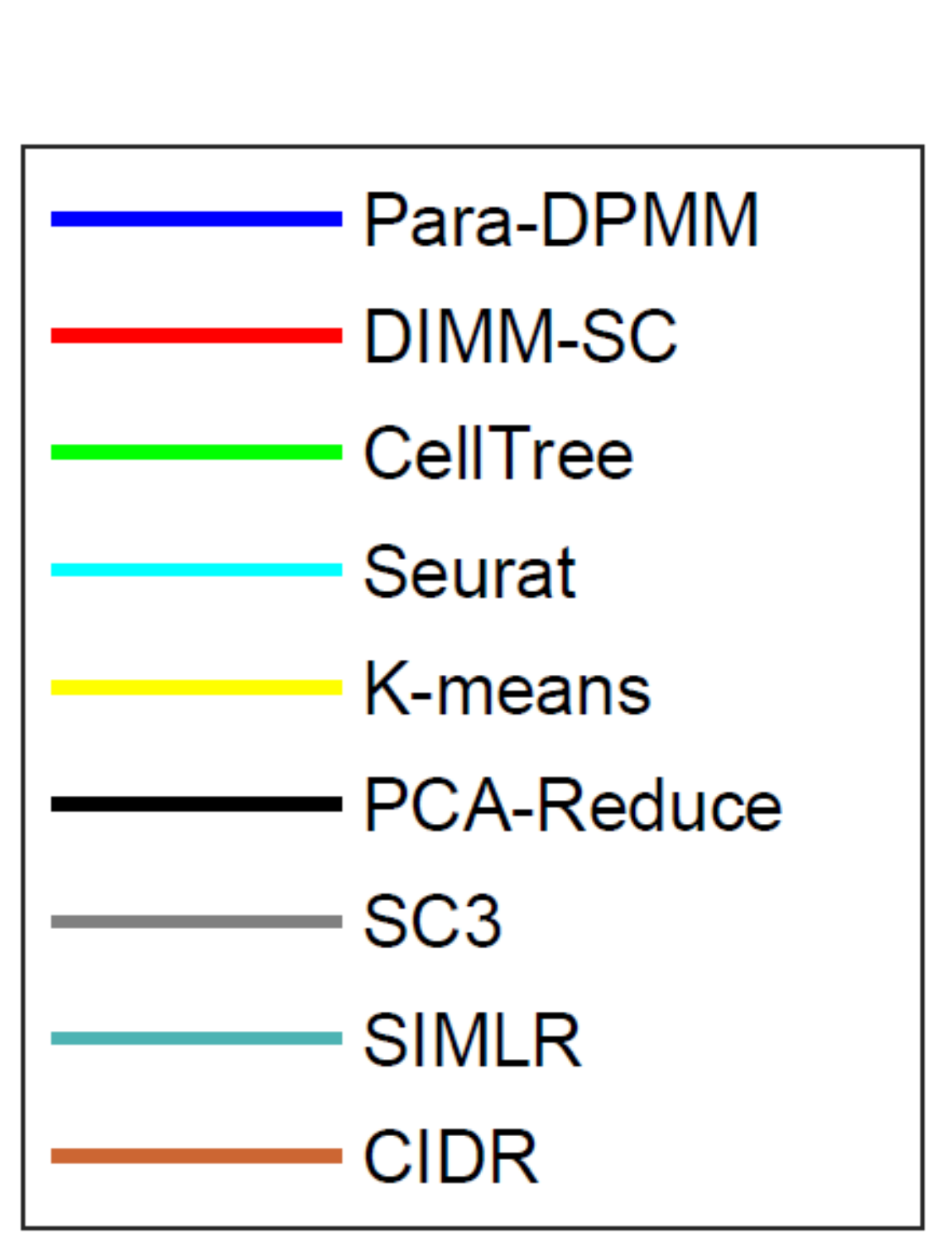}
		\label{Fig9_4}
	}
	\caption{(a) Performance (ARI) with respect to different number of genes on S-Set. (b) Performance with respect to different number of cells on S-Set. (c) Performance with respect to different number of genes on L-Set. (d) Performance with respect to different number of cells on L-Set.}
	\label{Fig9}
\end{figure*}

We further explored the relationship of model performance with different number of genes and cells. Results are presented in Fig. \ref{Fig9}. For the small scale setting, the performance slightly increased with gene number (Fig. \ref{Fig9_1}), as the cell clusters are more distinguishable with the added information. This result shows Para-DPMM's ability to handle the increasing dimensionality in data, as posterior inference of multinomial model only involves multiplying one dimension at a time and naturally circumvents the high dimensionality challenge. The DIMM-SC model achieved good performance with number of genes less than 1000. The Seurat algorithm performed better with the increase of the number of genes. Its clustering is based on embedding cells to graphs and analyzing the cliques formed. Increasing the number of genes made the edge weight more accurate. The performance of other comparison methods is not significantly influenced by number of genes. For large scale setting, the performance of Para-DPMM remained stable (Fig. (\ref{Fig9_3}) and Fig. (\ref{Fig9_4})). The performance slightly improved when more genes were involved, as more UMI counts are accumulated in the process and clusters becomes more distinguishable.

\begin{figure*}[ht]
	\centering
	\begin{minipage}{1\textwidth}
		\centering
		\subfloat[]
		{
			\includegraphics[scale=.45]{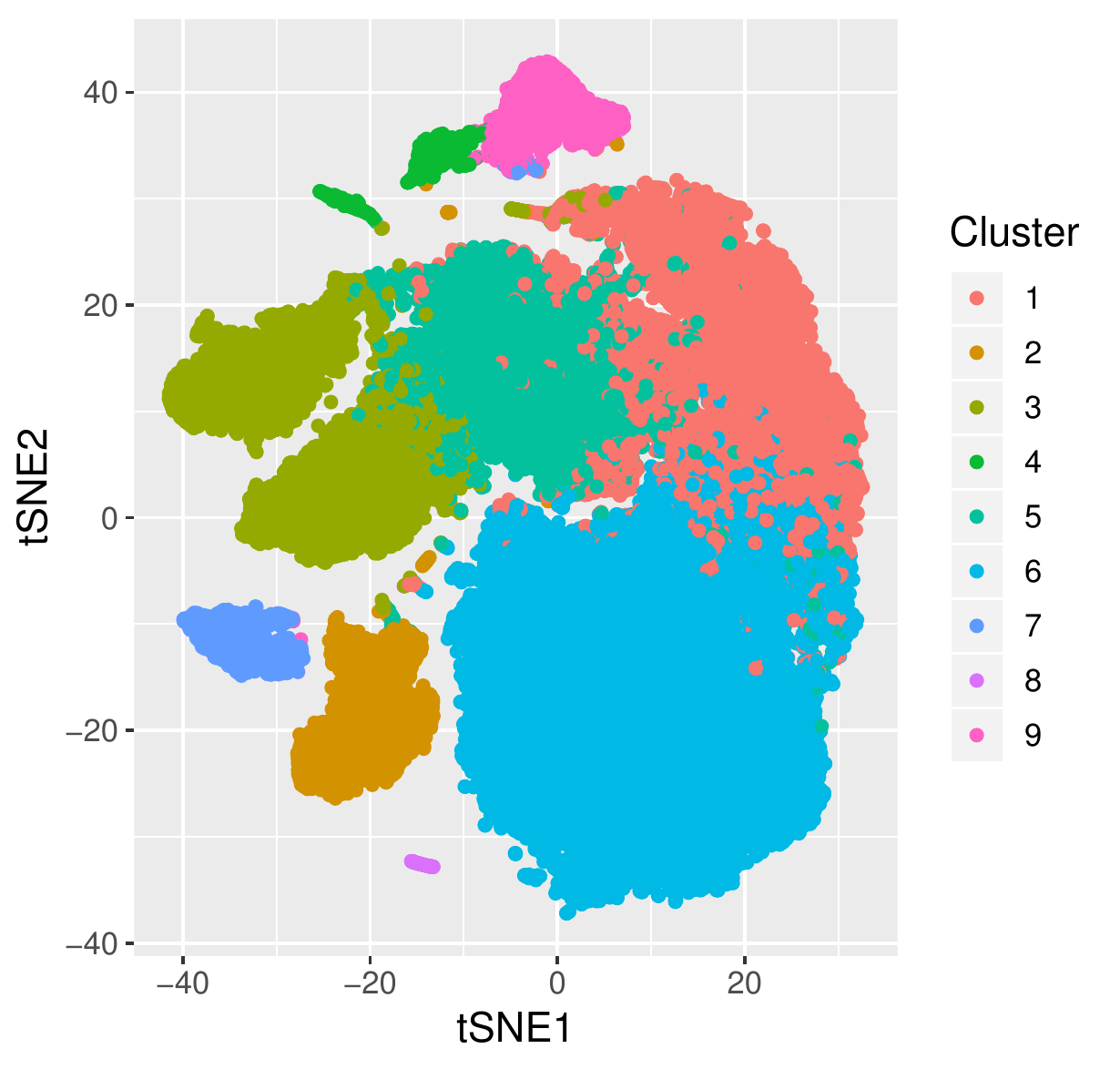}
			\label{Fig11_11}
		}
		\subfloat[]
		{
			\includegraphics[scale=.45]{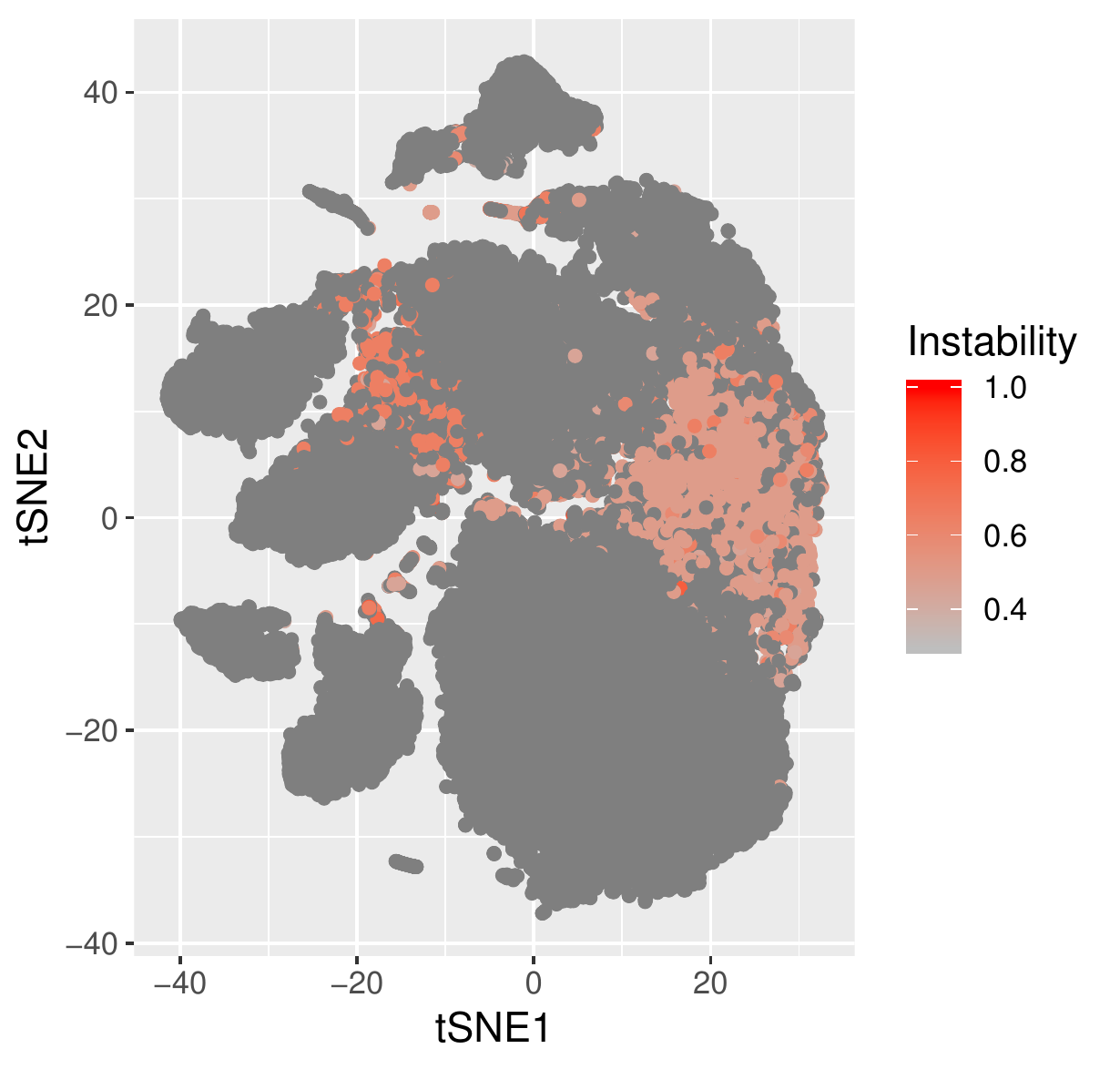}
			\label{Fig11_12}
		}
	\end{minipage}%
	
	\begin{minipage}{1\textwidth}
		\centering
		\subfloat[]
		{
			\includegraphics[scale=.2]{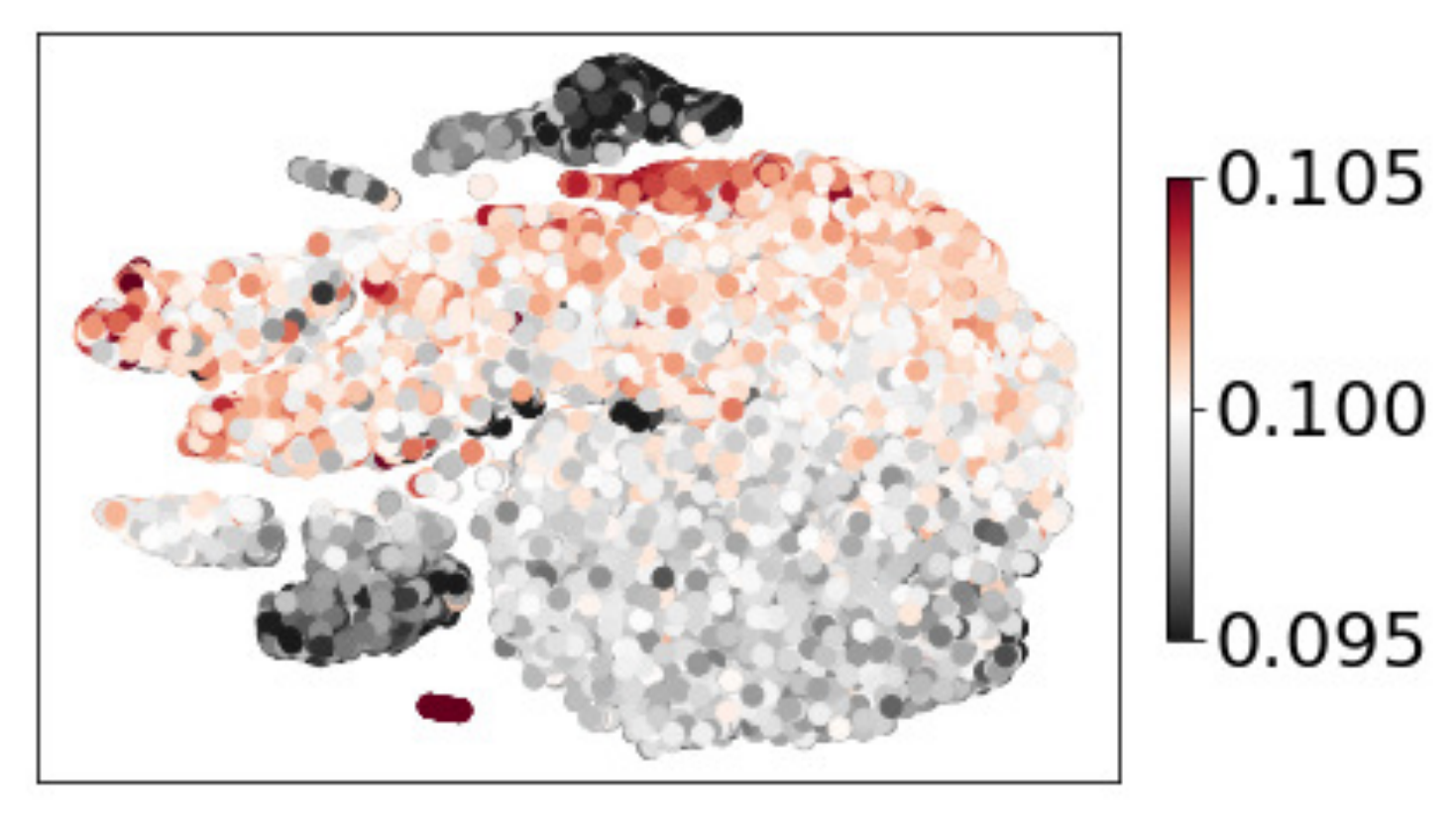}
			\label{Fig11_1}
		}
		\subfloat[]
		{
			\includegraphics[scale=.2]{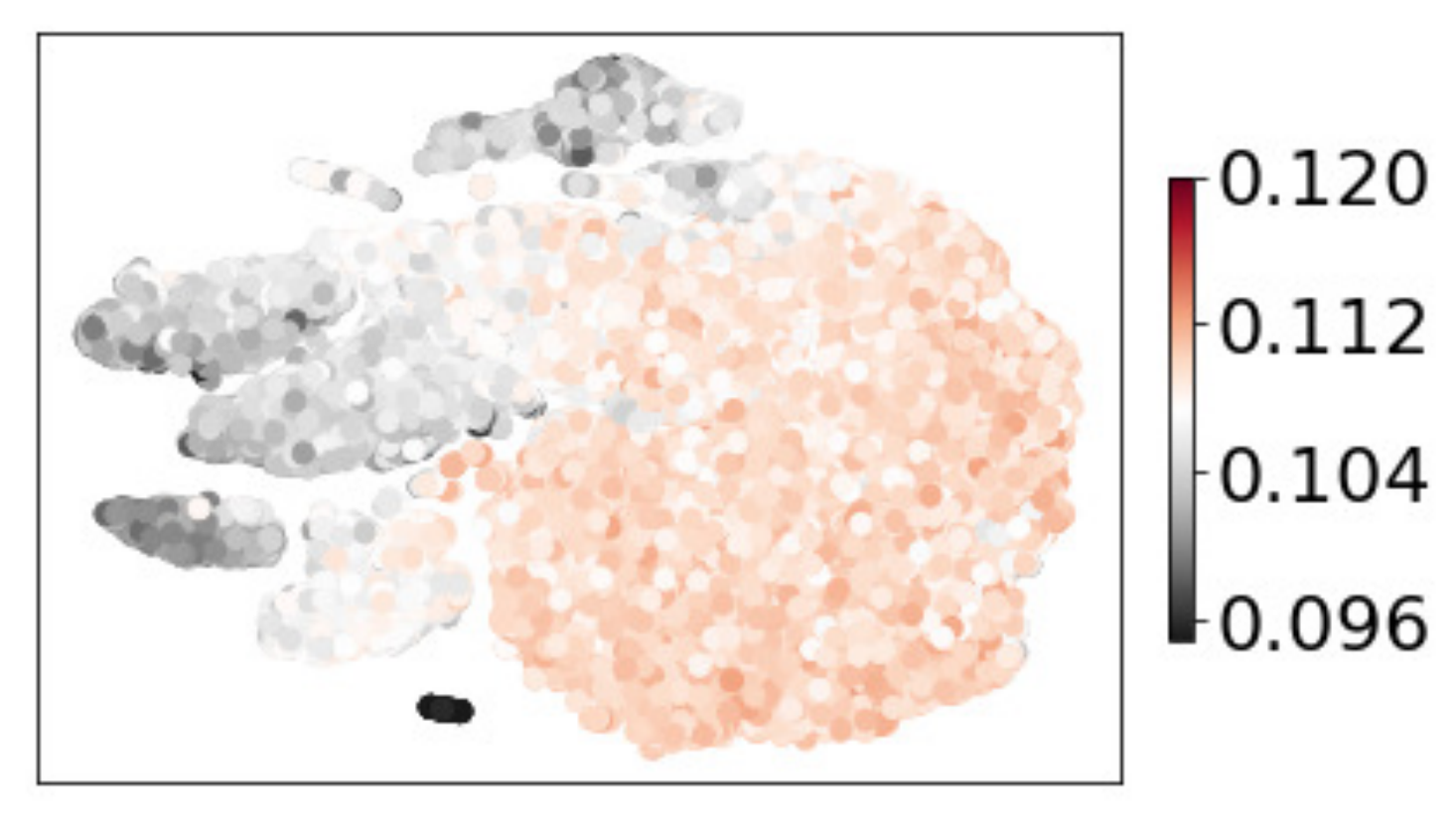}
			\label{Fig11_2}
		}
		\subfloat[]
		{
			\includegraphics[scale=.2]{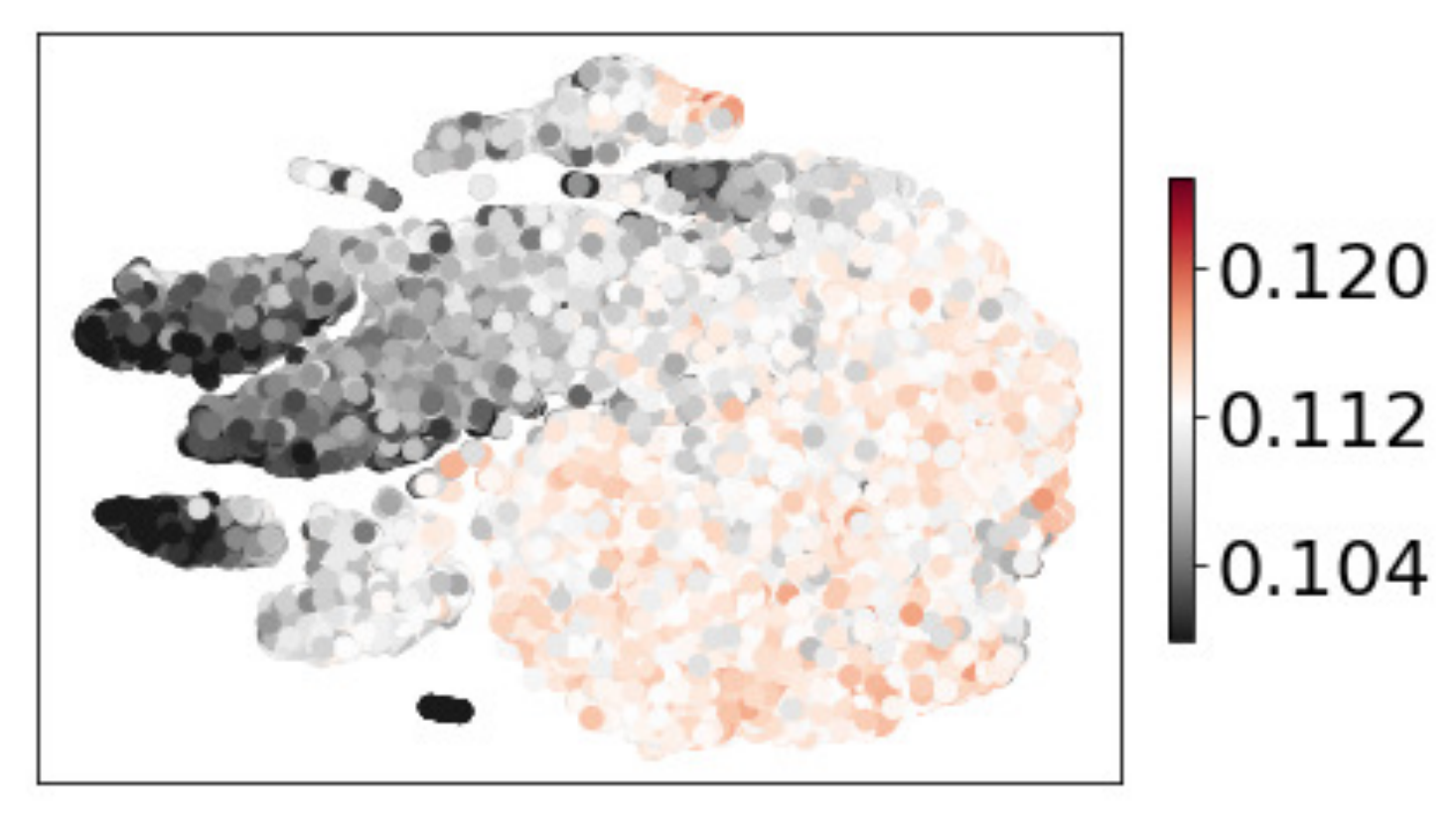}
			\label{Fig11_3}
		}
		\subfloat[]
		{
			\includegraphics[scale=.2]{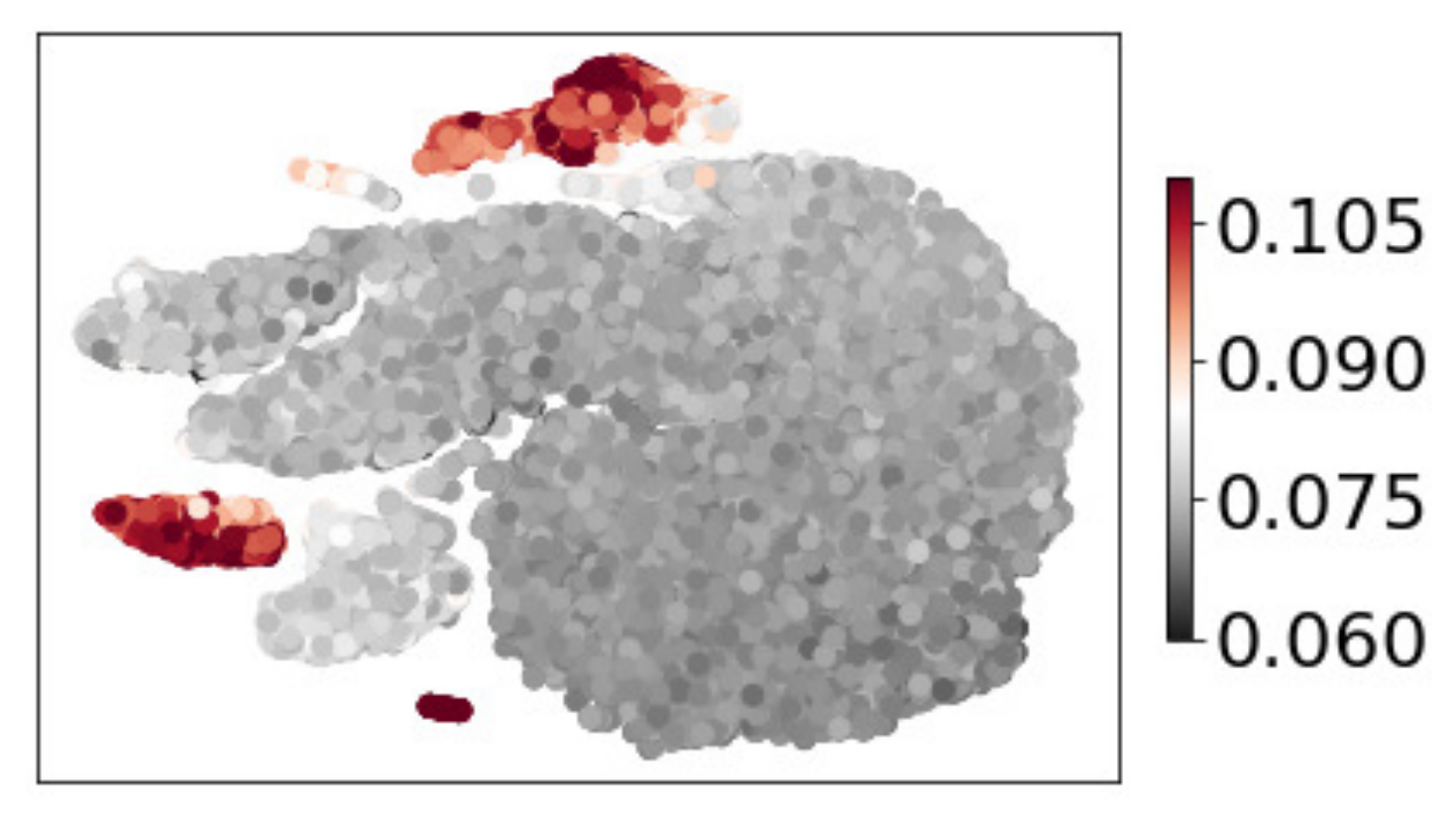}
			\label{Fig11_4}
		}
		\subfloat[]
		{
			\includegraphics[scale=.2]{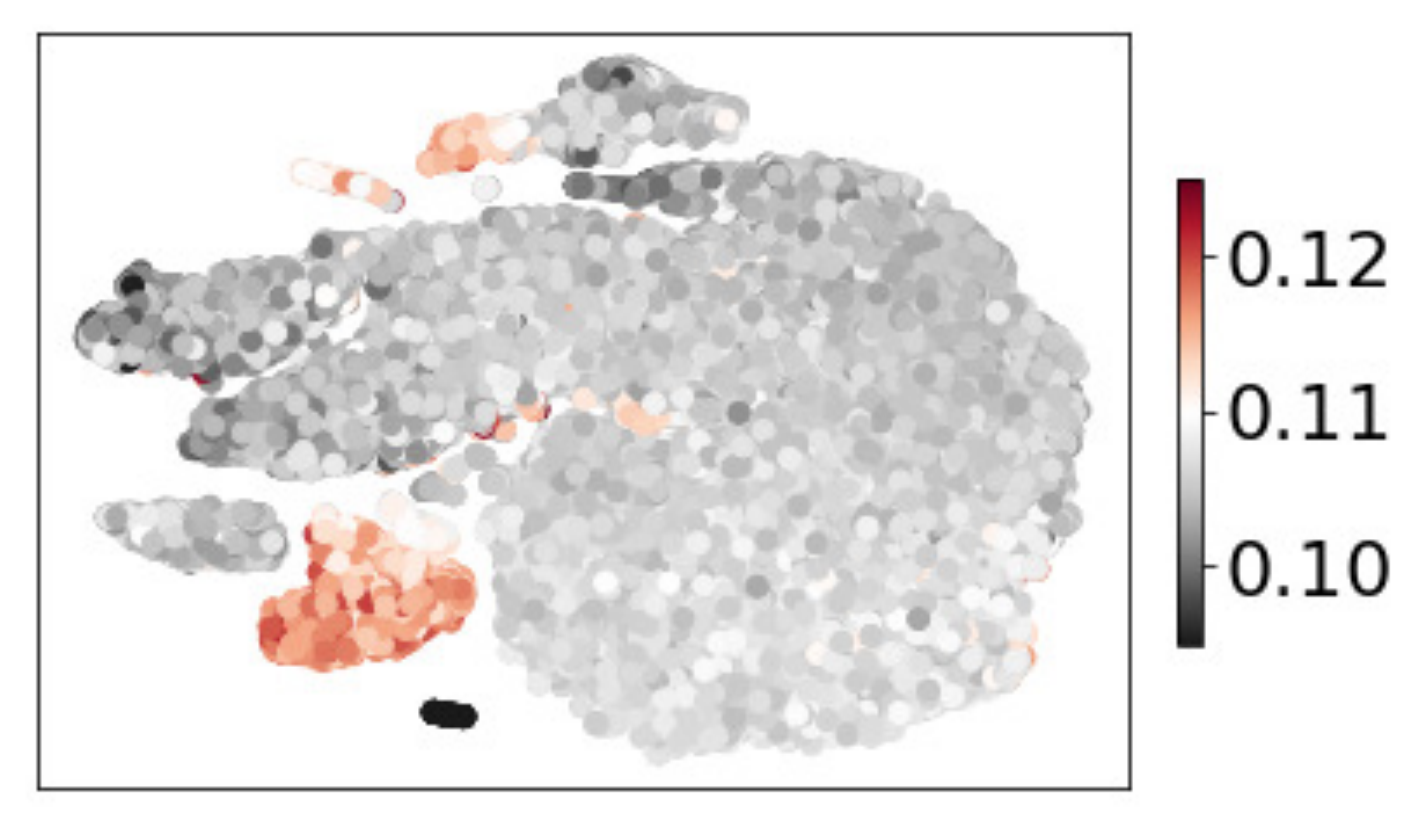}
			\label{Fig11_5}
		}
	\end{minipage}%
	
	\begin{minipage}{1.0\textwidth}
		\centering
		\subfloat[]
		{
			\includegraphics[scale=.2]{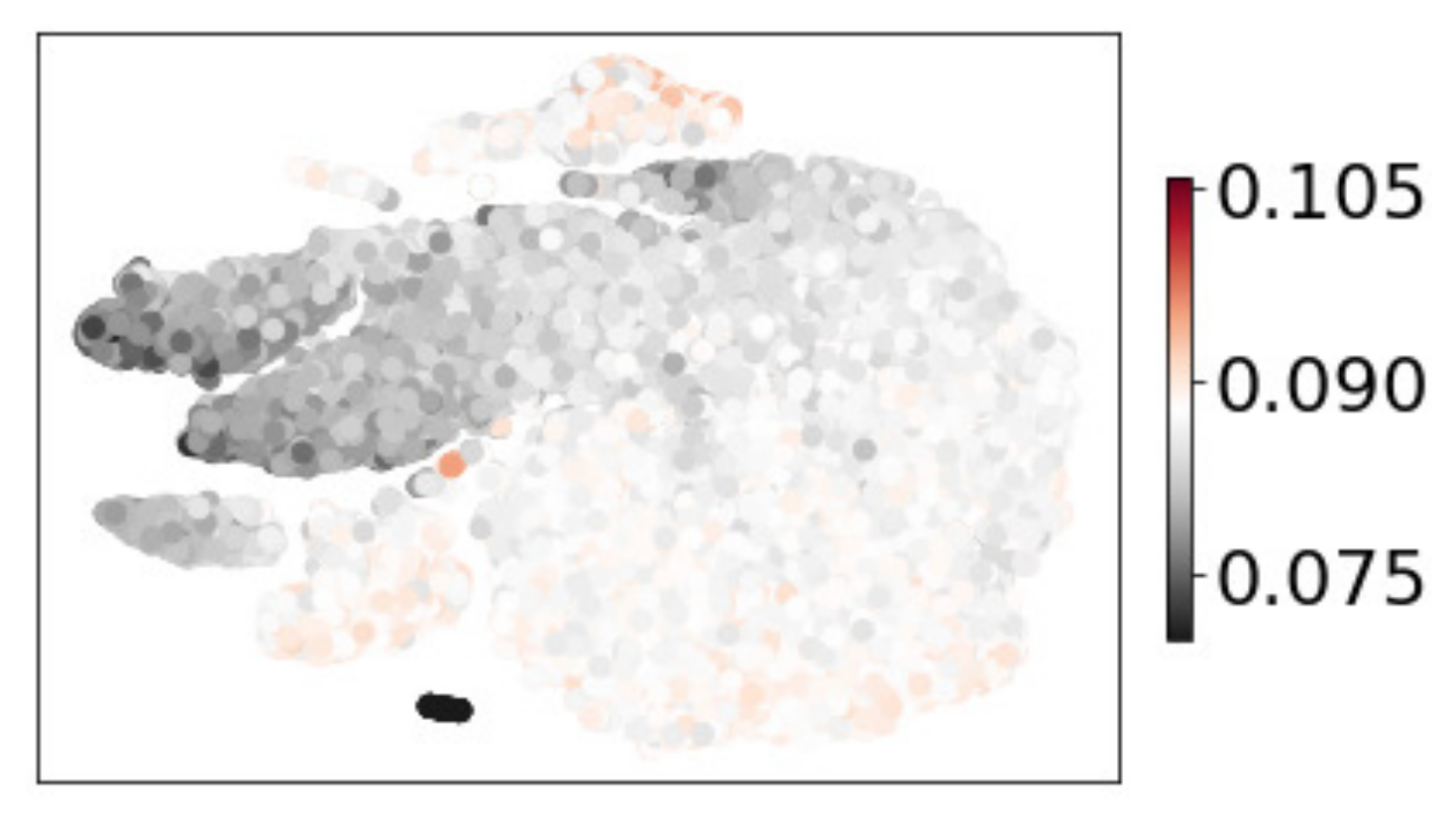}
			\label{Fig11_6}
		}
		\subfloat[]
		{
			\includegraphics[scale=.2]{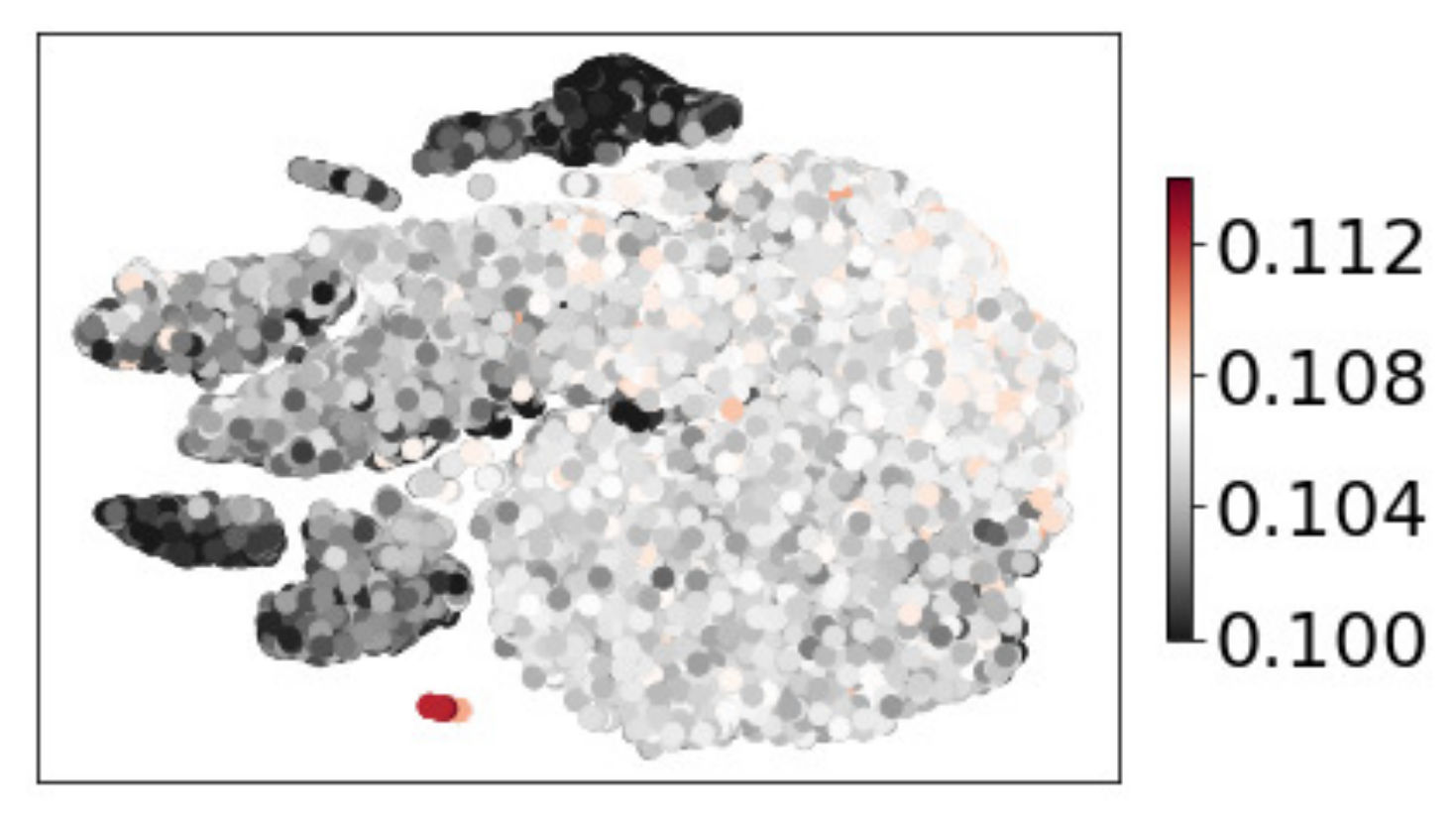}
			\label{Fig11_7}
		}
		\subfloat[]
		{
			\includegraphics[scale=.2]{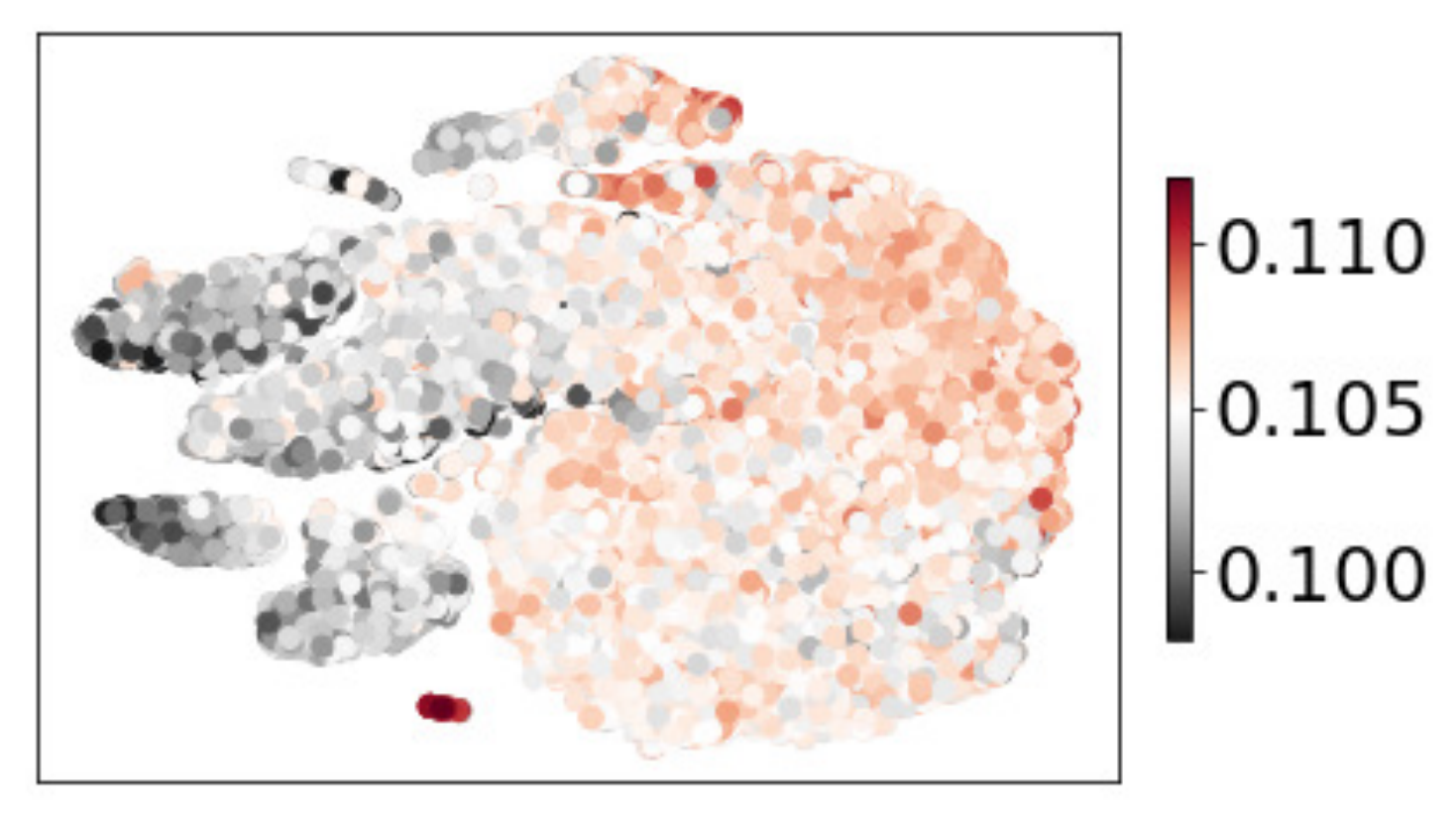}
			\label{Fig11_8}
		}
		\subfloat[]
		{
			\includegraphics[scale=.2]{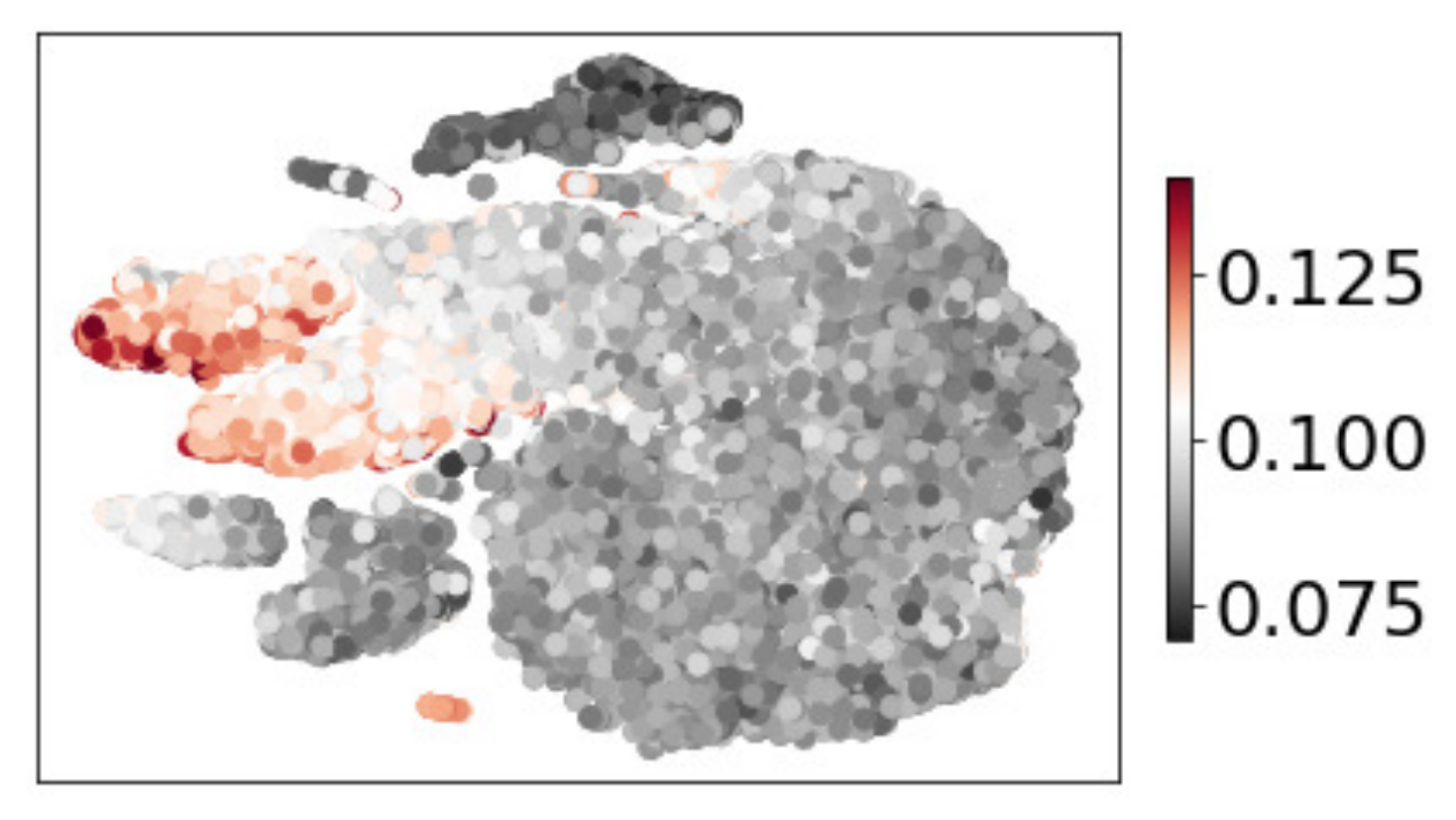}
			\label{Fig11_9}
		}
		\subfloat[]
		{
			\includegraphics[scale=.2]{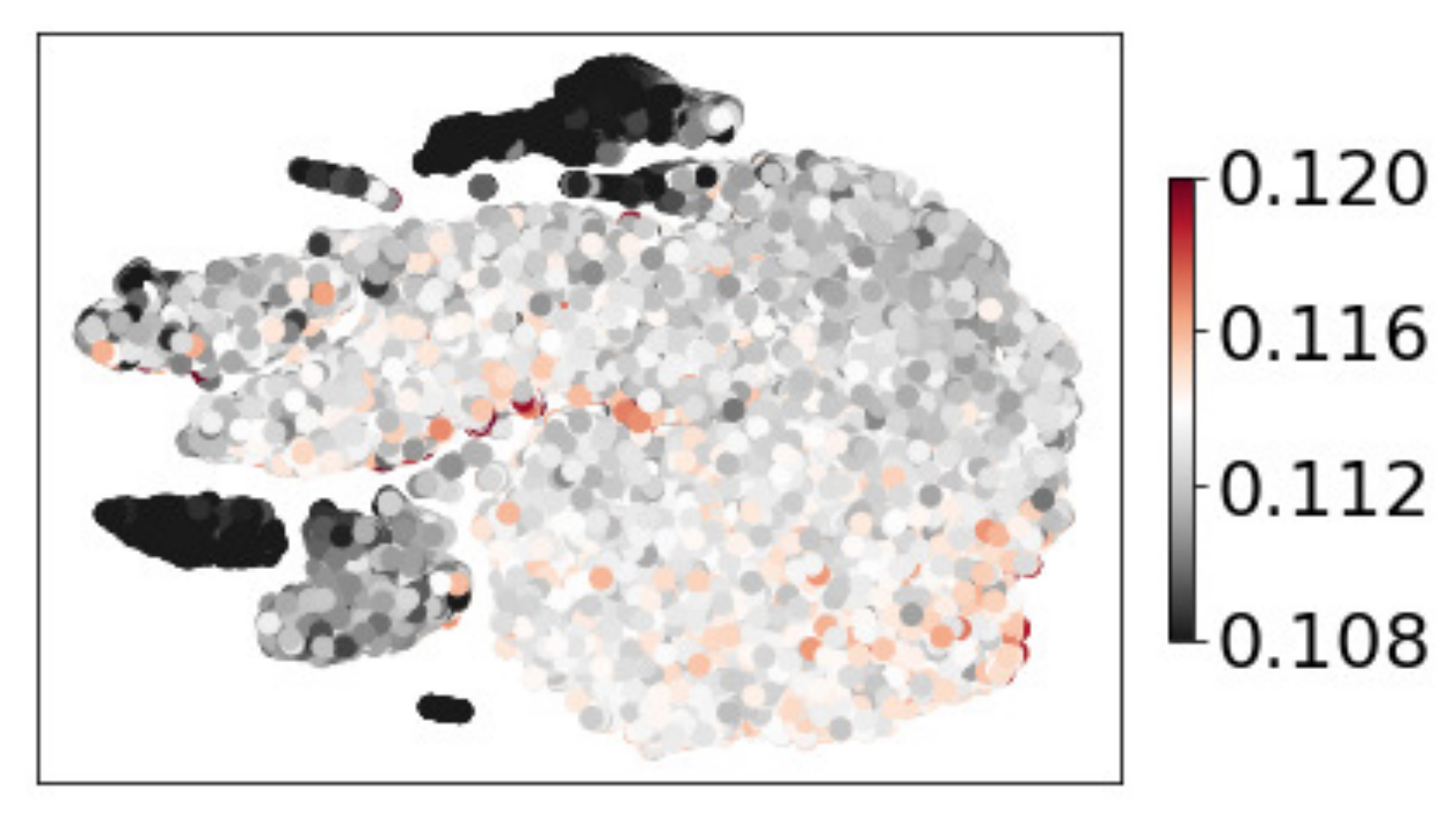}
			\label{Fig11_10}
		}
	\end{minipage}
	\caption{(a) t-SNE visualization of Para-DPMM clustering on Fresh PBMC 68K dataset; (b) stability of the clustering result; (c) CD4+/CD25+ regulatory T cell distribution; (d) CD4+/CD45ra+/CD25- naive T cell distribution; (e) CD8+/CD45ra+ naive cytotoxic T cell distribution; (f) CD14+ monocytes distribution; (g) CD19+ B cell distribution; (h) CD34+ cell distribution; (i) CD4+ helper T cell distribution; (j) CD4+/CD45ro+ memory T cell distribution; (k) CD56+ Natural Killer cell distribution; (l) CD8+ cytotoxic T cell distribution}
	\label{Fig11}
\end{figure*}

\section{Analysis of Fresh PBMC 68K Dataset} In order to demonstrate our model's ability to deal with real world large datasets, in this case study we applied Para-DPMM to a publicly available fresh PBMC 68K dataset\footnote{\label{note1}Publicly available on \url{https://support.10xgenomics.com/single-cell-gene-expression/datasets}}. The dataset is composed of 68K freshly processed peripheral blood mononuclear cells obtained from one donor. Samples are divided between T cells($>80\%$), NK cells($\sim 6\%$), B cells($\sim 6\%$) and myeloid cells($\sim 7\%$). Clustering analysis on the data reveals proportion of each cell types, identifies cell types with similar transcriptome profiles, finds finer grained subtypes in existing categories and discovers rare cell populations.

The results of the Para-DPMM clustering can be seen in Fig. (\ref{Fig11_11}). Our model divided the data points into 9 clusters, a result close to the 10 clusters identified with human expert knowledge (\citep{Zheng2017}). The clustering is in accordance with the boundaries of clusters visualized in the t-SNE plot. To test the stability of the clustering we repeated the process 50 times and measured the probability of each cell being assigned to different clusters. As illustrated in Fig. (\ref{Fig11_12}), the clusters were quite stable, though there was some uncertainty on the intersection regions of cluster 1 with 6 and cluster 3 with 5. We also tested the influence of hyper parameter $\alpha$ on the clustering result, and found different values of $\alpha$ had little effect on the clustering when ranging from 0.1 to 1. The reason for this robustness lies in the relative strength of prior (compared to likelihood) in determining posterior cluster distribution. Given the high dimensionality (number of genes) of the dataset, the likelihood dominates the posterior distribution in the sampling process and the small difference caused by different $\alpha$ in the prior distribution is negligible.

\begin{table*}[htbp]
	\centering
	\caption{Performance Comparison on Pairwise PBMC Cell Types}
	\begin{center}
		\scalebox{0.73}{
			\begin{tabular}{l|rrr|rrr|rrr}
				\toprule
				& \multicolumn{3}{c|}{CD4+CD45ro+/CD34+} & \multicolumn{3}{c|}{CD8+/CD4+CD45ra+CD25-} & \multicolumn{3}{c}{CD56+/CD4+CD25+} \\
				\midrule
				& \multicolumn{1}{c}{ARI} & \multicolumn{1}{c}{RI} & \multicolumn{1}{c}{HI} & \multicolumn{1}{c}{ARI} & \multicolumn{1}{c}{RI} & \multicolumn{1}{c}{HI} & \multicolumn{1}{c}{ARI} & \multicolumn{1}{c}{RI} & \multicolumn{1}{c}{HI} \\
				\midrule
				Para-DPMM & \textbf{0.706 $\pm$ 0.037} & \textbf{0.853 $\pm$ 0.019} & \textbf{0.706 $\pm$ 0.037} & \textbf{0.750 $\pm$ 0.035} & \textbf{0.875 $\pm$ 0.018} & \textbf{0.750 $\pm$ 0.035} & \textbf{0.990 $\pm$ 0.004} & \textbf{0.995 $\pm$ 0.002} & \textbf{0.990 $\pm$ 0.004} \\
				DIMM-SC & 0.672 $\pm$ 0.042 & 0.836 $\pm$ 0.021 & 0.672 $\pm$ 0.042 & 0.562 $\pm$ 0.048 & 0.781 $\pm$ 0.024 & 0.562 $\pm$ 0.048 & 0.971 $\pm$ 0.007 & 0.985 $\pm$ 0.003 & 0.971 $\pm$ 0.007 \\
				CellTree & 0.250 $\pm$ 0.031 & 0.625 $\pm$ 0.016 & 0.250 $\pm$ 0.031 & 0.161 $\pm$ 0.034 & 0.580 $\pm$ 0.017 & 0.161 $\pm$ 0.034 & 0.782 $\pm$ 0.038 & 0.891 $\pm$ 0.019 & 0.782 $\pm$ 0.038 \\
				Seurat & 0.432 $\pm$ 0.048 & 0.716 $\pm$ 0.024 & 0.432 $\pm$ 0.048 & 0.286 $\pm$ 0.012 & 0.643 $\pm$ 0.006 & 0.286 $\pm$ 0.012 & 0.581 $\pm$ 0.054 & 0.790 $\pm$ 0.027 & 0.581 $\pm$ 0.054 \\
				PCA-Reduce & 0.621 $\pm$ 0.040 & 0.811 $\pm$ 0.020 & 0.621 $\pm$ 0.040 & 0.459 $\pm$ 0.038 & 0.729 $\pm$ 0.019 & 0.459 $\pm$ 0.038 & 0.528 $\pm$ 0.032 & 0.764 $\pm$ 0.016 & 0.528 $\pm$ 0.032 \\
				K-Means & 0.202 $\pm$ 0.010 & 0.601 $\pm$ 0.005 & 0.202 $\pm$ 0.010 & 0.143 $\pm$ 0.008 & 0.572 $\pm$ 0.004 & 0.143 $\pm$ 0.008 & 0.746 $\pm$ 0.034 & 0.873 $\pm$ 0.017 & 0.746 $\pm$ 0.034 \\
				SC3 & \textbf{0.695 $\pm$ 0.026} & \textbf{0.847 $\pm$ 0.013} & \textbf{0.695 $\pm$ 0.026} & 0.709 $\pm$ 0.016 & 0.855 $\pm$ 0.008 & 0.709 $\pm$ 0.016 & 0.980 $\pm$ 0.004 & 0.991 $\pm$ 0.002 & 0.980 $\pm$ 0.004 \\
				SIMLR & 0.465 $\pm$ 0.034 & 0.761 $\pm$ 0.017 & 0.465 $\pm$ 0.034 & 0.376 $\pm$ 0.017 & 0.721 $\pm$ 0.008 & 0.376 $\pm$ 0.017 & 0.726 $\pm$ 0.026 & 0.878 $\pm$ 0.013 & 0.726 $\pm$ 0.026 \\
				CIDR & 0.684 $\pm$ 0.014 & 0.859 $\pm$ 0.007 & 0.684 $\pm$ 0.014 & 0.430 $\pm$ 0.012 & 0.745 $\pm$ 0.006 & 0.430 $\pm$ 0.012 & 0.823 $\pm$ 0.011 & 0.921 $\pm$ 0.005 & 0.823 $\pm$ 0.011 \\
				\bottomrule
			\end{tabular}%
		}
	\end{center}
	\label{tab:addlabe4}%
\end{table*}%


Since there is no available ground truth cell labeling for this dataset to obtain detailed knowledge about the specific cell types which compose the clusters, we resorted to 10 purified cell populations\footref{note1} of the cell types that were previously identified in this dataset using human expert knowledge. The cell type's gene expression profile was obtained by averaging the profiles of each purified population. The cell type assignment was based on the covariance between profiles of the cell types and samples. The distribution of each cell type is visualized in Fig. (\ref{Fig11_1}) to Fig. (\ref{Fig11_10}). CD14+ monocytes, CD19+ B cells and CD56+ NK cells were easily separated from other cell types. On the other hand, we observed a significant overlap of  CD4+/CD45+/CD25- naive T cell, CD8+/CD45ra+ naive cytotoxic T cells and CD4+/CD45+ memory cells on the t-SNE plot.

 These cell type distributions easily explain certain clusters, more specifically clusters 2, 3 and 7, which are composed mostly of  CD19+ B cells,  CD56+ NK cells and CD14+ monocytes respectively. Other clusters are composed of multiple cell types. Cluster 6 is a combination of CD4+/CD45+/CD25- naive T cells and CD8+/CD45ra+ naive cytotoxic T cells, clusters 1 and 5 also contain a significant amount of these cell types while being mainly composed of CD4+/CD25+ regulatory T cells.

We found that three pairs of cells were largely overlapping in the clusters, namely CD4+/CD45ro+ memory T with CD34+ cells, CD8+ cytotoxic T with CD4+/CD45ra+/CD25- naive T cells and CD56+ Natural Killer with CD4+/CD25+ regulatory T cells. We further tested our model's ability to distinguish these three pairs of cells. 2,000 cells from each category were randomly selected and clustered based on the 16,000 genes with top expression variation. Results are presented in Table \ref{tab:addlabe4}. The performance of SC3 was comparable to Para-DPMM for the CD4+CD45ro+/CD34+ pair. Para-DPMM achieved better performance than all comparison methods in the other two pairs. We found it was significantly easier to distinguish between CD56+ Natural Killer and CD4+/CD25+ regulatory T cells than the other two pairs.

\section{Applicable Scenario Analysis} \label{domain}

The Para-DPMM model should be applied to datasets created with UMI based techniques. In UMI labeling based systems, the UMI counts are independent of transcript length and is suitable to model with Multinomial distribution. As illustrated in \citep{Islam2013} and \citep{Phipson2017}, earlier non-UMI based techniques introduced bias during the cDNA amplification phase, the resulting expression matrix is correlated with transcript length and normalizations used in RPKM and FPKM are necessary. For these datasets, clustering methods based on continuous similarity measures such as Seurat, SC3 and PCA-Reduce are more appropriate choices.

\begin{figure}[ht]
	\centering
	\includegraphics[scale=0.5]{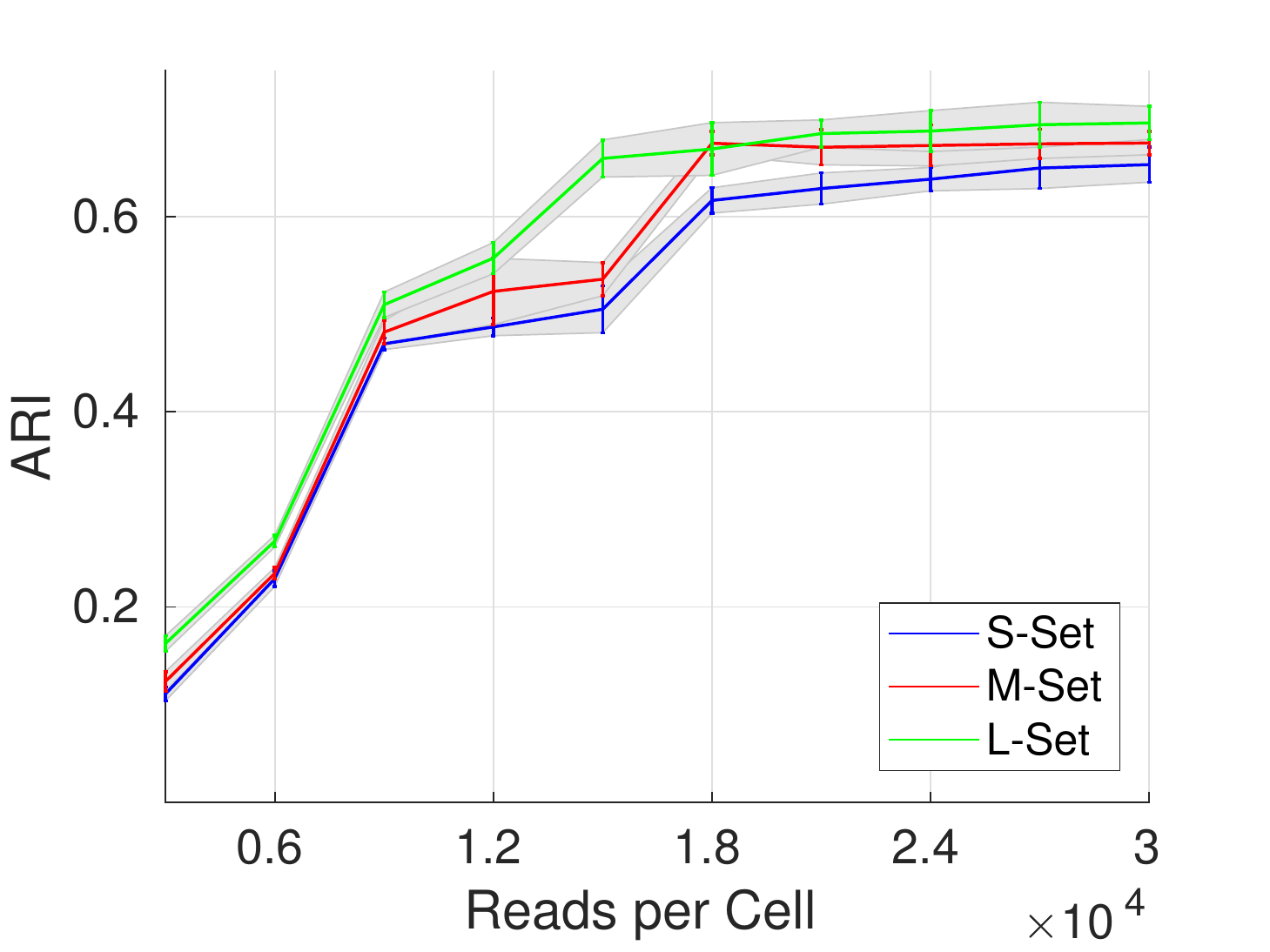}
	\caption{Influence of Sequencing Depth on Model Performance}
	\label{Fig14}
\end{figure}

Current droplet-based single cell sequencing techniques has the drop out phenomenon, where not all transcriptome information is captured during the cell reads. This results in a sparser expression matrix when the sequencing depth is not deep enough. To test the robustness of Para-DPMM regarding to varying sequencing depth, we measured the model performance on different data scales (S-Set, M-Set and L-Set) with sequencing depth ranging from 3,000 to 30,000 reads per cell. As shown in Fig. \ref{Fig14}, the model performance is highly correlated with sequencing depth when reads per cell is less than 10,000, and performance is stable after sequencing depth reaches 18,000 reads per cell. The recommended minimum sequencing depth for 10X platform is 50,000 reads per cell (\citep{Baran_2017}), which lies well inside the model's robust region. 

\section{Scalability Analysis on Parallel Computing Clusters}

In this section, we analyze the scalability of the model. Para-DPMM was implemented on a HPC cluster built with the BeeGFS system, the model uses the OpenMP framework and is able to run in parallel on multiple cores in one node. We tested the model's scalability with up to 32 cores. Further improvement on parallelization is possible if the model is extended with the MPI framework, which is not in the scope of this paper. We requested 64GB RAM for all experiment settings.


\begin{figure*}[ht]
	\centering
	\subfloat[]
	{
		\includegraphics[scale=.33]{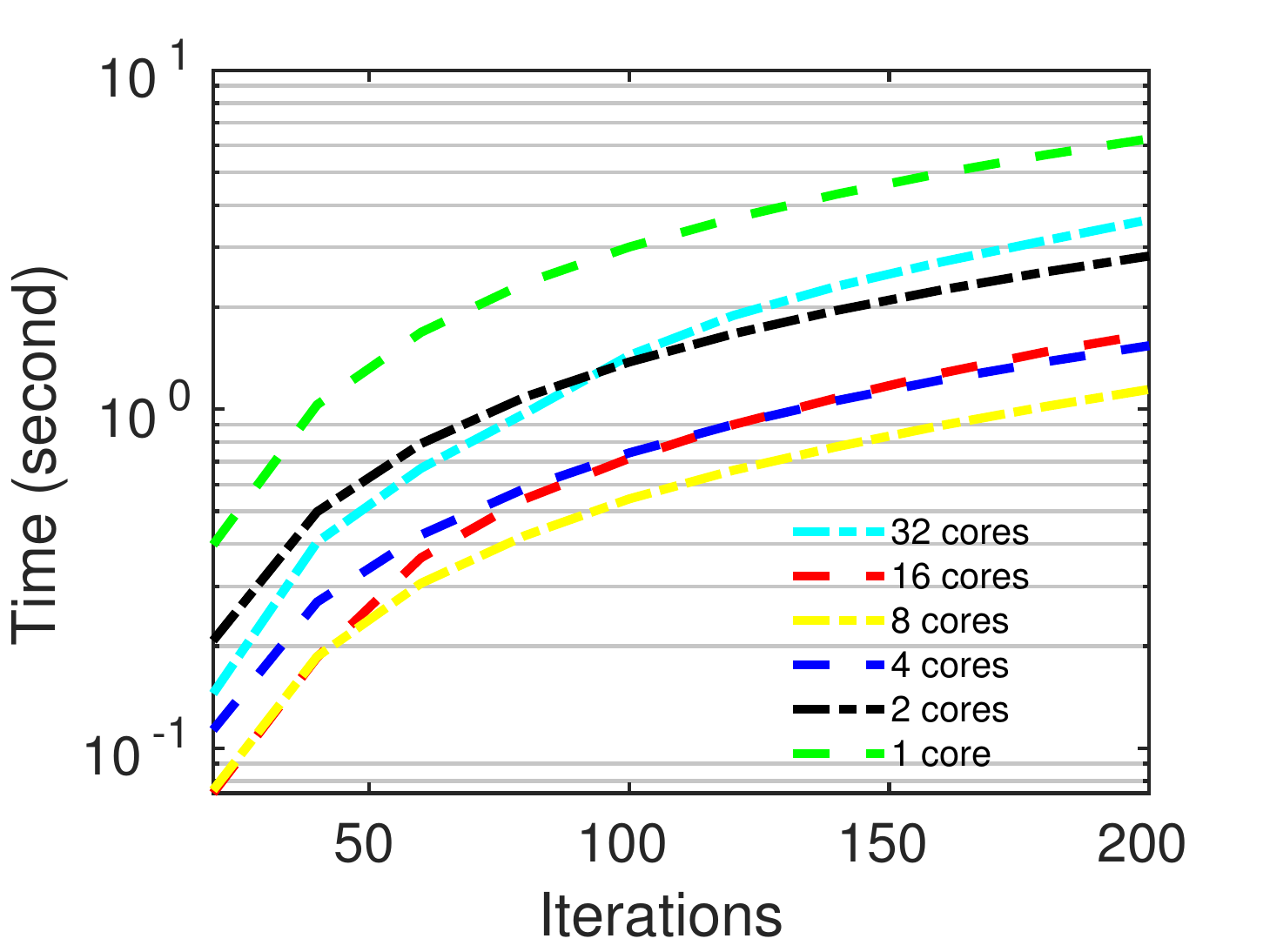}
		\label{Fig12_1}
	}
	\subfloat[]
	{
		\includegraphics[scale=.33]{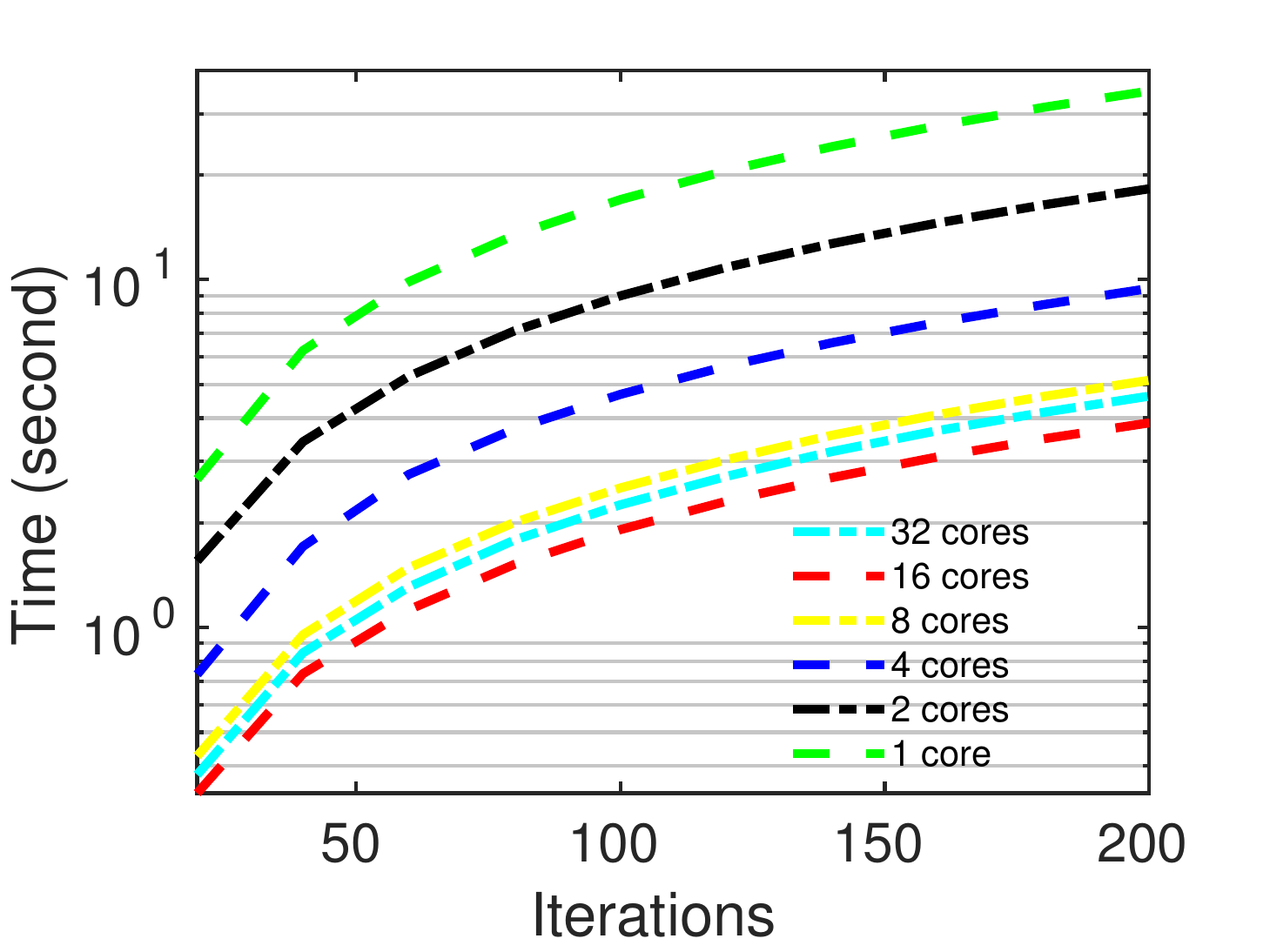}
		\label{Fig12_2}
	}
	\subfloat[]
	{
		\includegraphics[scale=.33]{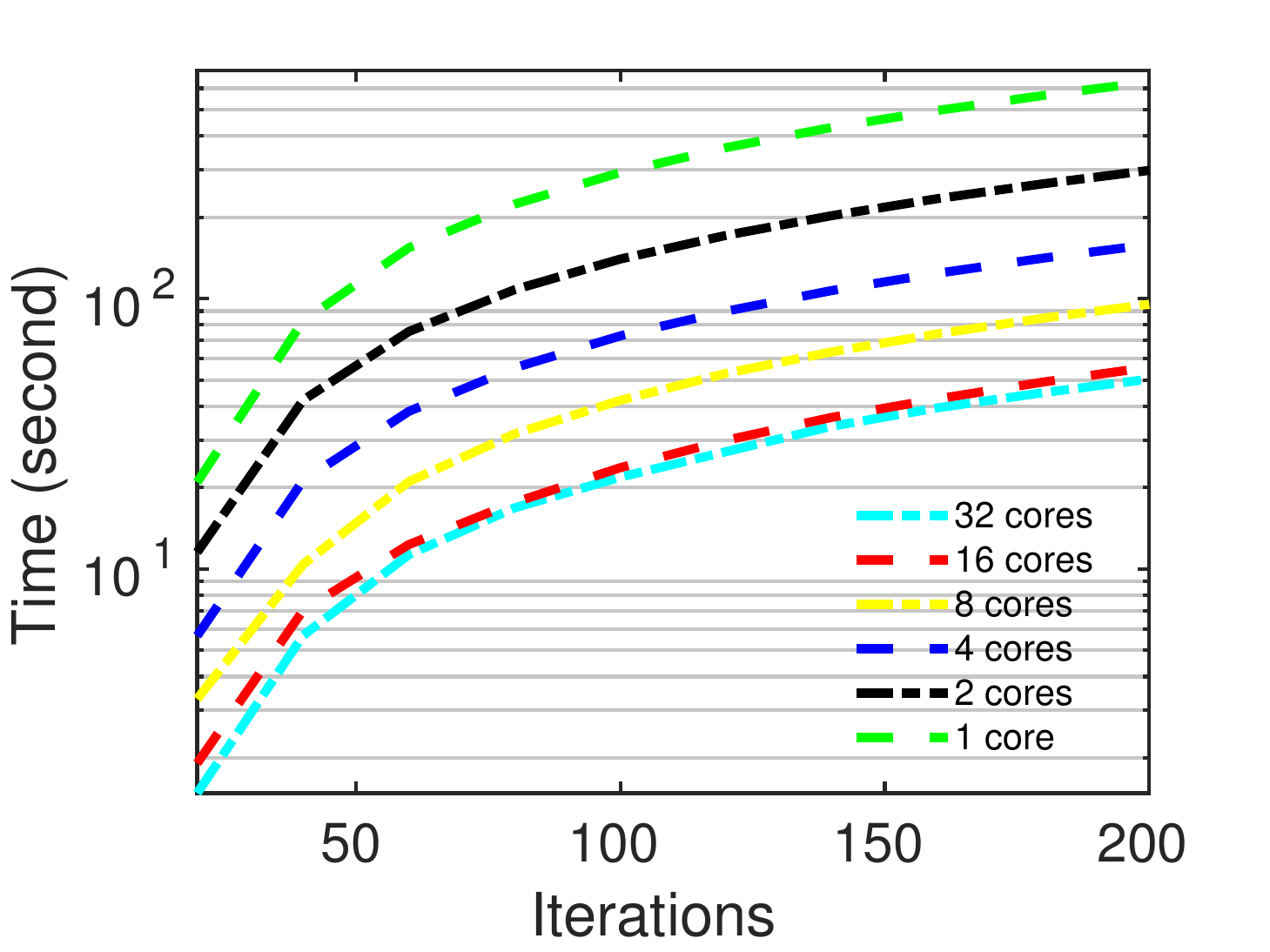}
		\label{Fig12_3}
	}
	\caption{(a) Comparison of Computing Time on S-Set; (b) Comparison of Computing Time on L-Set; (c) Comparison of Computing Time on PBMC 68K dataset.}
	\label{Fig12}
\end{figure*}

We recorded the model's computing time on variating number of cores for different dataset sizes, results are shown in Fig. (\ref{Fig12}). The trade off between the gain and cost of parallelization is clearly exemplified on the small dataset (S-Set, shown in Fig. (\ref{Fig12_1})), where fastest computing speed was achieved with 8 computing cores, after which computing became slower as the number of cores further increased. The cost of parallelization came from coordination between different threads, including parallel tasks creation, I/O of the shared memory and communications between threads, which eventually offsets the gains. Fig. (\ref{Fig12_1}) demonstrates it is not necessary to use more than 8 cores for training on the small dataset. The strength of parallelized implementation becomes evident when dealing with large scaled datasets, such as the PBMC 68K data. As shown in Fig. (\ref{Fig12_3}), the computing speed is approximately 12 times faster when using 32 cores compared to a single core. The computing time is initially inversely proportional to the number of cores, and then gradually converge to constant time.

Based on Amdahl's Law
\begin{equation} \label{eq:18}
\text{Speed Up}=\frac{1}{\frac{P}{N}+S}
\end{equation}
where $P$ denotes the parallelized portion in the code, $N$ denotes number of cores and $S=1-P$ denotes the serial portion in the code, the parallelization ratio of the model implementation is as high as 91\%.

 \begin{table*}[!ht]
	\centering
	\caption{Computing Speed Comparison of Different Models}
	\scalebox{1.0}{
		\begin{tabular}{l|r|r|r|r|r|r|r|r|r}
			\toprule
			& \multicolumn{1}{c|}{Para-DPMM} & \multicolumn{1}{c|}{DIMM-SC} & \multicolumn{1}{c|}{CellTree} & \multicolumn{1}{c|}{Seurat} &\multicolumn{1}{c|}{PCA-reduce} & \multicolumn{1}{c|}{K-means} & \multicolumn{1}{c|}{SC3} & \multicolumn{1}{c|}{SIMLR} & \multicolumn{1}{c}{CIDR} \\
			\midrule
			S-Set & 1.14 sec & 33.10 sec & 1.82 sec & 28.46 sec & 5.56 sec & 5.30 sec & 3.11 min & 9.26 min & 7.09 sec \\
			M-Set & 2.16 sec & 4.77 min & 3.06 sec & 1.23 min & 2.07 min & 20.12 sec & 5.21 min & 1.28 hrs & 54.39 sec \\
			L-Set & 3.88 sec & 16.98 min & 6.41 sec & 2.48 min & 11.10 min & 48.95 sec & 8.43 min & 6.65 hrs & 6.77 min \\
			\bottomrule
		\end{tabular}%
	}
	\label{tab:addlabe2}%
\end{table*}%

We also compared other models' computing speed\protect\footnotemark with Para-DPMM (Table \ref{tab:addlabe2}). For fairness, the measurements include only running time and exclude time for data I/O and dimension reduction (in Seurat). All models were run on 8 cores and towards convergence. Para-DPMM and CellTree are significantly faster than other comparison methods. Para-DPMM is about 30\% faster than CellTree on small data setting and 40\% on large settings.

\footnotetext{Please note the computing time is significantly affected by factors at software engineering level. This comparison should only serve as guidance for real world applications, and not to be used for inferring algorithm complexity.}

\section{Discussion} As shown in the experiments, the Para-DPMM model scales well with different dataset size (Table \ref{label1}) and with variating data dimensionality (Fig. \ref{Fig9}). This scalability and versatility enables its possible wide application on real world genomic systems. Clustering analysis on the fresh PBMC dataset (Fig. \ref{Fig11_11}) identified cells with similar transcriptome profiles and helped uncover finer grained heterogeneous structures for each cell type. As illustrated in the applicable scenario analysis (section \ref{domain}), the model should only be applied to UMI-based datasets. 

 To cope with the large scaled single cell transcriptomic datasets, the model's inference process is highly parallelized and ready for applications in large computing clusters. This parallelization is achieved by explicitly instantiating the cluster parameters of the model and makes data points conditionally independent of each other. While the model can potentially utilize as many computing cores as the number of data points, 32 cores are generally enough for current large datasets (Fig. \ref{Fig12_3}).

 The split-merge mechanism is adopted in the model to significantly improve convergence and optimality of the result. The integrated split-merge process is formed with two independent MCMC chains which generates high acceptance ratio for both split and merge moves. We performed detailed comparison with current widely used methods, and Para-DPMM model simultaneously achieved significant improvements on both clustering accuracy and computing speed. The model's performance increases with higher dimensionality of the data, and it automatically infers number of clusters from the dataset without using prior knowledge. 

Several extensions of the Para-DPMM model are possible. For single cell datasets created from heterogeneous sources (e.g. PBMC cells from multiple individuals), the model could be extended to include hierarchical processes to discover fine grained substructures in the clusters. Given the availability of purified cell populations, the clustering accuracy could be further improved with semi-supervised guidance. We will explore these possible extensions in the near future.

\bibliographystyle{ieeetr}

\bibliography{Para_DPMM}

\appendix
\gdef\thesection{Appendix: } 
\section{Derivation of Acceptance Ratio in Split/Merge MCMC Sampler}

According to the generative procedure of DPMM, $\frac{p(S^{*})}{p(S)}$ can be decomposed as

\begin{equation} \label{eq:29}
\left.\begin{aligned}
\frac{p(S^{*})}{p(S)}=\frac{p(\vec{\pi}_{*}, \vec{\theta}_{*}, \vec{c}_{*}, \vec{x}_{*})}{p(\vec{\pi}, \vec{\theta}, \vec{c}, \vec{x})}=\frac{p(\vec{\pi}_{*})}{p(\vec{\pi})}\frac{p(\vec{c}_{*}|\vec{\pi}_{*})}{p(\vec{c}|\vec{\pi})}\frac{p(\vec{\theta}_{*}|\lambda)}{p(\vec{\theta}|\lambda)}\frac{p(\vec{x}|\vec{c}_{*},\vec{\theta}_{*})}{p(\vec{x}|\vec{c},\vec{\theta})}
\end{aligned}\right.
\end{equation}

When we perform Gibbs sampling, we are directly forming the posterior probabilities, and we have

\begin{equation} \label{eq:10}
p(\vec{\pi})p(\vec{c}|\vec{\pi})=p(\vec{c})p(\vec{\pi}|\vec{c})=\frac{\alpha^{K}\prod_{k}\Gamma(n_{k})}{\Gamma(\alpha+N)}\frac{\Gamma(\alpha+N)}{\Gamma(\alpha)\prod_{k}\Gamma(n_{k})}\pi_{K+1}^{\alpha}\prod_{k}\pi_{k}^{n_{k}-1}=\frac{\alpha^{K}}{\Gamma(\alpha)}\pi_{K+1}^{\alpha}\prod_{k}\pi_{k}^{n_{k}-1}
\end{equation}

The cells in each cluster are modeled through the multinomial distribution, the cluster parameter $\vec{\theta}_{k}$ for cluster k are the set of burstiness probability $\theta_{k,u}$ for each gene $u \in \{1,2,..,V\}$  following Dirichlet distribution with hyper parameter $\lambda$. We have

\begin{equation} \label{eq:11}
p(\vec{\theta}|\lambda)=\frac{\Gamma(\lambda V)^{K}}{\Gamma(\lambda)^{V K}}\prod_{k=1}^{K}\prod_{u=1}^{V}\theta_{k,u}^{\lambda-1}
\end{equation}

\begin{equation} \label{eq:13}
\begin{split}
p(\vec{x}|\vec{c},\vec{\theta})=\prod_{k=1}^{K}\prod_{i \in \{k\}}Mult(\vec{x}_{i}|\vec{\theta}_{k})=\prod_{k=1}^{K}\prod_{i\in \{k\}}\prod_{u=1}^{V}\theta_{k,u}^{x_{i}^{u}}
\end{split}
\end{equation}

\begin{equation} \label{eq:12}
\begin{split}
p(\vec{\theta}|\lambda)p(\vec{x}|\vec{c},\vec{\theta})=\frac{\Gamma(\lambda V)^{K}}{\Gamma(\lambda)^{V K}}\Big(\prod_{k=1}^{K}\prod_{u=1}^{V}\theta_{k,u}^{\lambda-1}\Big)\Big(\prod_{k=1}^{K}\prod_{i\in \{k\}}\prod_{u=1}^{V}\theta_{k,u}^{x_{i}^{u}}\Big)
\end{split}
\end{equation}

The acceptance ratio $p(S^{*},S)$ can be readily computed after substituting Eq. (\ref{eq:10}) and Eq. (\ref{eq:12}) into Eq. (\ref{eq:29}). Clusters other than the splitted cluster k are not changed and terms regrading these clusters are canceled out. We have

\begin{equation} \label{eq:14}
\frac{p(\vec{\pi}_{*})p(\vec{c}_{*}|\vec{\pi}_{*})}{p(\vec{\pi})p(\vec{c}|\vec{\pi})}=\alpha \frac{\pi_{k_{0}}^{\bar{n}_{k_{0}}-1}\pi_{k_{1}}^{\bar{n}_{k_{1}}-1}}{\pi_{k}^{n_{k}-1}}
\end{equation}

where $k_{0}$ and $k_{1}$ are the two subclusters formed after splitting cluster $k$.

Using the uniform prior $\lambda$ for all $\vec{\theta}_{k}$, we have

\begin{equation} \label{eq:16}
\frac{p(\vec{\theta}_{*}|\lambda)p(\vec{x}_{*}|\vec{c}_{*},\vec{\theta}_{*})}{p(\vec{\theta}|\lambda)p(\vec{x}|\vec{c},\vec{\theta})}=\frac{\Gamma(\lambda V)}{\Gamma(\lambda)^{V}}\frac{\prod_{u=1}^{V}\theta_{k_{0},u}^{\lambda-1}\prod_{u=1}^{V}\theta_{k_{1},u}^{\lambda-1}}{\prod_{u=1}^{V}\theta_{k,u}^{\lambda-1}}\times\frac{\Big(\prod_{i\in \{k_{0}\}}\prod_{u=1}^{V}\theta_{k_{0},u}^{x_{i}^{u}}\Big)\Big(\prod_{i\in \{k_{1}\}}\prod_{u=1}^{V}\theta_{k_{1},u}^{x_{i}^{u}}\Big)}{\Big(\prod_{i\in \{k\}}\prod_{u=1}^{V}\theta_{k,u}^{x_{i}^{u}}\Big)}
\end{equation}

Combining Eq. \ref{eq:14} and Eq. \ref{eq:16}, the complete form of $\frac{p(S^{*})}{p(S)}$ is

\begin{equation} \label{eq:18}
\frac{p(S^{*})}{p(S)}=\alpha \frac{\pi_{k_{0}}^{\bar{n}_{k_{0}}-1}\pi_{k_{1}}^{\bar{n}_{k_{1}}-1}}{\pi_{k}^{n_{k}-1}}\frac{\Gamma(\lambda V)}{\Gamma(\lambda)^{V}}\frac{\prod_{u=1}^{V}\theta_{k_{0},u}^{\lambda-1}\prod_{u=1}^{V}\theta_{k_{1},u}^{\lambda-1}}{\prod_{u=1}^{V}\theta_{k,u}^{\lambda-1}}\times\frac{\Big(\prod_{i\in \{k_{0}\}}\prod_{u=1}^{V}\theta_{k_{0},u}^{x_{i}^{u}}\Big)\Big(\prod_{i\in \{k_{1}\}}\prod_{u=1}^{V}\theta_{k_{1},u}^{x_{i}^{u}}\Big)}{\Big(\prod_{i\in \{k\}}\prod_{u=1}^{V}\theta_{k,u}^{x_{i}^{u}}\Big)}
\end{equation}

\end{document}